%% file: main.tex
\documentclass[sigconf]{acmart}

\usepackage{enumitem}
\usepackage{booktabs,multirow}

\usepackage{pgfplots}
\pgfplotsset{compat=1.18}
\definecolor{mutgnblue}{HTML}{245A81}
\definecolor{sharingblue}{HTML}{91B4C7}
\definecolor{normamber}{HTML}{D6A064}

\newtheorem{proposition}{Proposition}
\newtheorem{corollary}{Corollary}

\AtBeginDocument{%
  }

\begin{document}

\title{MuTGN: A Multilinear Tensor-based Autoencoder for Anomaly Detection in Temporal Graphs}

\author{Maryam Rahmani}
\affiliation{
  \institution{Univ. Littoral Côte d'Opale \\ LISIC - UR 4491}
  \country{F-62219 Longuenesse, France}}
\additionalaffiliation{
  \institution{ICAM, Lille Campus, 6 rue Auber, 59016 Lille Cedex, France}
}
\email{maryam.rahmani@icam.fr}

\author{Claire Guilloteau}
\affiliation{
  \institution{Univ. Littoral Côte d'Opale \\ LISIC - UR 4491}
  \country{F-62219 Longuenesse, France}}
\email{claire.guilloteau@univ-littoral.fr}

\author{Esteban Bautista}
\affiliation{
  \institution{Univ. Littoral Côte d'Opale \\ LISIC - UR 4491}
  \country{F-62219 Longuenesse, France}}
\email{esteban.bautista@univ-littoral.fr}

\renewcommand{\shortauthors}{Rahmani et al.}

\begin{abstract}

Anomaly detection in temporal graphs is an important challenge with numerous applications. In recent years, a large number of methods for detecting edge, node, and graph-level anomalies have been developed. Yet, most methods target anomalies that manifest at individual timestamps, overlooking window-level anomalies: anomalies that may arise from an abnormal temporal ordering of graph snapshots. While some tensor-based methods have been proposed for detecting abnormal windows, existing solutions struggle with the sparsity of real-world temporal graphs and neglect the nodal attributes. To address these limitations, this work introduces MuTGN, a novel tensor-based autoencoder architecture for window anomaly detection in attributed temporal graphs. MuTGN leverages multilinear algebra to map entire windows of an attributed temporal graph into a compact latent core tensor from which both the evolving graph topology and time-varying node attributes can be reconstructed. Anomalies are then detected by training MuTGN on normal data and identifying windows with high reconstruction errors. In particular, MuTGN is (a) \emph{temporally and structurally aware}: it can assess whether any segment of an attributed temporal graph is dynamically and/or topologically abnormal; (b) \emph{aggregation-free}: it is designed to operate directly on highly sparse temporal graphs without requiring time aggregates to densify graph snapshots; (c) \emph{efficient}: for a window containing $N$ nodes, $K$ timestamps, $M_A$ observed time-stamped edges, and $M_X$ observed time-node feature vectors, its inference time complexity is $\mathcal{O}(K + N + M_A + M_X)$; and (d) \emph{effective}: it significantly outperforms state-of-the-art alternatives in detecting a broad range of injected anomalies and successfully detects a real-world Internet outage.

\end{abstract}

\ccsdesc[500]{Computing methodologies~Learning latent representations}
\ccsdesc[500]{Computing methodologies~Anomaly detection}
\ccsdesc[300]{Information systems~Data stream mining}
\ccsdesc[300]{Computing methodologies~Neural networks}

\keywords{Anomaly detection, Temporal graphs, Tensor learning, Autoencoders}

\received{20 February 2007}
\received[revised]{12 March 2009}
\received[accepted]{5 June 2009}

\maketitle

\input{Sources/Introduction}

\input{Sources/Related_works}

\input{Sources/Proposed_method}

\input{Sources/Numerical_experiments}

\section{Conclusion}

In this work, we introduced MuTGN, a multilinear autoencoder for window anomaly detection in attributed temporal graphs. MuTGN leverages feature-conditioned projections of the adjacency tensor to encode entire segments of an attributed temporal graph into compact latent core tensors, from which both the evolving topology and the node attributes can be reconstructed. Its design allows MuTGN to operate directly on highly sparse temporal graphs without requiring temporal aggregation or dense materialization of the input tensors, while achieving an inference time complexity that, for fixed architectural dimensions, is linear in the size of the input data. Our extensive experimental evaluation demonstrated that MuTGN is able to detect a broad range of window anomalies and to outperform existing methods in many scenarios, particularly those involving abnormal temporal dynamics. Furthermore, MuTGN demonstrated its practical relevance beyond injected anomalies by successfully identifying a documented Internet outage in a large-scale temporal graph representing the Internet at the autonomous-system level. The next steps involve improving MuTGN's sensitivity to certain types of anomalies, such as sparsification and rewiring, which remain challenging. Another promising direction is to move beyond window-level detection toward anomaly localization, identifying the subset of an anomalous window that contributes the most to its anomaly score.

\section*{Acknowledgments}
This work is funded in part by the ANR (French National Agency of Research) through the CPJ contract of Esteban Bautista (ANR-23-CPJ1-0048-01) and the France 2030 Data2Laws (ANR-24-EXEN-0002) grants. We would like to thank Justin Loye and Romain Fontugne for providing access to the Internet outge data.

\bibliographystyle{ACM-Reference-Format}
\bibliography{biblio}

\appendix

\input{Sources/Appendix_complexity_analysis}

\input{Sources/Appendix_attack_descriptions}

\input{Sources/Appendix_negative_sampling}
\input{Sources/Appendix_model_selection}

\input{Sources/Appendix_synthetic_data}

\input{Sources/Appendix_performance_tables}

\input{Sources/Appendix_ablation_tables}

\end{document}

%% file: Sources/Introduction.tex
\section{Introduction}

Network traffic, road trips, bank transfers, and email communications are examples of data that can be naturally modeled as attributed temporal graphs. For instance, in network traffic, nodes can represent computers, time-stamped edges can represent packets, and node attributes can describe device information such as resource utilization and open ports. Similarly, in bank transactions, nodes can represent bank accounts, time-stamped edges can represent transfers, and node attributes can model information such as account balances. Detecting anomalies in these networks is an important problem that has attracted considerable attention in recent years. In particular, numerous methods have been proposed to detect anomalous edges \cite{bhatia.2022.midas, chang.2021.ffade, liu.2023.taddy, eswaran.2018.sedanspot, liu.2024.anomalyllm, liu.2025.comedy}, nodes \cite{bansal.2022.dsedn, jiao.2021.geabs, lee.2024.slade, chakraborty.2025.flex, liu.2024.stripe}, graph snapshots \cite{yoon.2019.anomrank, guo.2022.dynanom, eswaran.2018.spotlight, huang.2020.lad, mcneil.2023.cadence, luo.2025.graph}, and specific substructures such as communities \cite{jiang.2022.tsad, tasnadi.2022.new} and subgraphs \cite{bautista.2024.mad}. Notably, some of these methods have been successfully applied in tasks such as network intrusion detection \cite{king.2023.edgetorrent}, Internet disruption monitoring \cite{justin.2026.madbgp}, and financial fraud detection \cite{khodabandehlou.2024.fifraud}.

\begin{figure*}
\includegraphics[width=\textwidth]{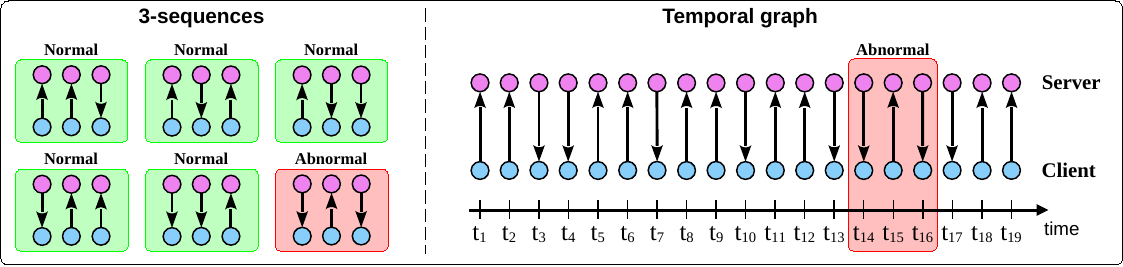}
\caption{
Illustration of a window anomaly. The right panel shows a temporal graph representing communications between a client and a server. The left panel shows the 3-sequence motifs obtained by sliding a window of length three. Frequent motifs are highlighted in green and a rare/abnormal motif is highlighted in red. The anomaly arises from the temporal ordering of the graph snapshots rather than from any individual snapshot.
}\label{fig:introduction-window_anomaly}
\end{figure*}

Despite these successes, most algorithms still overlook window-level anomalies: anomalies that can arise from an abnormal temporal ordering of graph snapshots\footnote{In this work a snapshot refers to a time-stamped graph}, even when these snapshots can be individually normal. In particular, most algorithms are not well suited to detecting such anomalies as they operate by flagging individual graphs from the sequence, at individual timestamps, yet window anomalies cannot be attributed to any particular graph from the sequence. Figure \ref{fig:introduction-window_anomaly} illustrates this phenomenon. The right panel shows a temporal graph representing traffic between a client and a server. At each timestamp, the graph contains either an outgoing or an incoming link, both of which occur frequently and can thus be considered normal. However, analyzing the traffic over sliding windows of length three reveals a window anomaly. The left panel shows the different 3-sequences obtained by sliding the window, with a stride of one, across the temporal graph. The motifs in green occur repeatedly and can be considered normal, while the motif in red occurs only once and can be deemed anomalous. Importantly, the three snapshots forming the anomalous window also participate in normal windows when the window is shifted one timestamp to the left or right. Thus, the window anomaly cannot be attributed to any particular graph from the sequence: it is a property of the three snapshots when considered collectively. This example therefore motivates the development of dedicated methods for window anomaly detection.

In the literature, a few methods for window anomaly detection have been proposed \cite{sun.2006.dta, koutra.2012.tensorsplat, leggas.2021.knowledge, fernandes.2021.wintended, xu.2019.swtf}. These approaches, in a nutshell, leverage tensor analysis to jointly characterize the structure and the dynamics of segments of a temporal graph, assuming that a normal segment possesses a low-rank structure and that anomalies disrupt such property. However, these tensor-based methods suffer from two main drawbacks. Firstly, they rely on low-rank tensor decompositions that often struggle with the high sparsity of many real-world temporal graphs, for which they often perform time aggregates as a means to densify the graph snapshots, yet such aggregation may destroy the signature of a window anomaly. Secondly, these approaches omit any available node attributes and thus are unable to detect anomalies that manifest in the attributes.

In order to address these limitations, we introduce MuTGN\footnote{Code available at \url{https://github.com/estbautista/MuTGN}.}, a novel temporal graph autoencoder architecture for window anomaly detection in attributed temporal graphs. MuTGN builds upon multilinear algebra to develop an encoder that jointly maps the adjacency and attribute tensors of each window into a compact latent core tensor, from which dedicated decoder heads, also based on multilinear operations, are able to reconstruct both tensors. This allows MuTGN to detect window anomalies affecting either the evolving graph topology or the time-stamped node attributes. In sum, our contributions are the following:

\begin{itemize}[leftmargin=*]
    \item \emph{Window anomalies in attributed temporal graphs}: MuTGN is capable of detecting both structural and dynamic anomalies that manifest in either the graph topology or the node attributes at the window level.
    \item \emph{A novel autoencoder architecture}: MuTGN is a novel autoencoder architecture based on multilinear algebra that goes beyond standard approaches for dynamic graph representation learning by embedding entire windows of attributed temporal graphs rather than individual edges, nodes, or graph snapshots.
    \item \emph{Aggregation-free detection}: MuTGN is designed to operate directly on highly sparse temporal graphs, without requiring temporal aggregation to densify graph snapshots.
    \item \emph{Efficiency}: For a window containing $N$ nodes, $K$ timestamps, $M_A$ observed time-stamped edges, and $M_X$ observed time-node feature vectors, the inference time complexity of MuTGN is $\mathcal{O}(K + N + M_A + M_X)$.
    \item \emph{Effectiveness}: MuTGN significantly outperforms state-of-the-art alternatives in an extensive evaluation involving 1 synthetic and 5 real-world datasets with 11 types of injected window anomalies. MuTGN also demonstrates its practical utility through the successful detection of a real-world Internet outage.
\end{itemize}

%% file: Sources/Related_works.tex
\section{Notations and Related Works}

\subsection{Notations}

Let $\mathbb{V}$ denote a set of vertices of size $N = |\mathbb{V}|$, $\mathbb{T} = \{0, 1, \dots, T - 1\} \subset \mathbb{N}$ refer to a discrete time-domain, $\mathbb{E} \subseteq \mathbb{V} \times \mathbb{V} \times \mathbb{T}$ be a set of time-stamped edges, and $\mathbb{P} \subset \mathbb{V} \times \mathbb{T}$ denote a set of time-stamped nodes. We define a discrete-time attributed temporal graph as the sixtuple $\mathcal{G}=(\mathbb{V}, \mathbb{T}, \mathbb{E}, \mathbb{P}, \ell, x)$, where $\ell : \mathbb{E} \to \mathbb{R}_{+}$ assigns weights to time-stamped edges and $x : \mathbb{P} \to \mathbb{R}^{F}$ assigns $F$-dimensional feature vectors to time-stamped nodes. We refer to a consecutive time interval $\mathbb{W} = \{t_0, t_0 + 1, \ldots, t_0 + K-1 \} \subseteq \mathbb{T}$ as a window of length $K$, and we let $\mathcal{G}_{\mathbb{W}} = (\mathbb{V}, \mathbb{W}, \mathbb{E}_{\mathbb{W}}, \mathbb{P}_{\mathbb{W}}, \ell_{\mathbb{W}}, x_{\mathbb{W}})$ denote the restriction of $\mathcal{G}$ to $\mathbb{W}$. We define the adjacency tensor of $\mathcal{G}$ as $\mathcal{A} \in \mathbb{R}^{N \times N \times T}$, where $\mathcal{A}[u, v, t] = \ell(u, v, t)$ if $(u,v,t) \in \mathbb{E}$ and zero otherwise. Moreover, we let $\mathcal{X} \in \mathbb{R}^{N \times T \times F}$ be the feature tensor of $\mathcal{G}$, in which $\mathcal{X}[u, t, f] = x(u, t)[f]$ if $(u, t) \in \mathbb{P}$ and zero otherwise. The adjacency and feature tensors of $\mathcal G_{\mathbb{W}}$ are defined as $\mathcal{A}_{\mathbb{W}} \in \mathbb {R}^{N \times N \times K}$ and $\mathcal{X}_{\mathbb{W}} \in \mathbb{R}^{N \times K \times F}$, respectively, where $\mathcal{A}_{\mathbb {W}}[u,v,k] = \mathcal{A}[u, v, t_0 + k]$ and $\mathcal{X}_{\mathbb{W}}[u, k, f] = \mathcal{X}[u, t_0 + k, f]$, for $k \in \{0, \ldots, K-1 \}$. Additionally, we let $M_A = |\mathbb{E}_{\mathbb{W}}|$ and $M_X = |\mathbb{P}_{\mathbb{W}}|$. Given a $P$-order tensor $\mathcal{Z} \in \mathbb{R}^{I_1 \times \cdots \times I_P}$ and a matrix $\mathbf{U} \in \mathbb{R}^{J \times I_n}$, the $n$-mode product of $\mathcal{Z}$ and $\mathbf{U}$, denoted $\mathcal{Y} = \mathcal{Z} \times_n \mathbf{U}$, where $\mathcal{Y} \in \mathbb{R}^{I_1 \times \dots \times I_{n-1} \times J \times I_{n+1} \times \dots \times I_P}$, is defined as $\mathcal{Y}[i_1, \dots, j, \dots, i_P] = \sum_{i_n = 1}^{I_n} \mathcal{Z}[i_1, \dots, i_n, \dots, i_P] \mathbf{U}[j, i_n]$. Given $P$ matrices $\mathbf{U}^{(p)} \in \mathbb{R}^{J_p \times I_p}$, $p = 1, \dots, P$, the multilinear product of $\mathcal{Z}$ and such collection of matrices is defined as $\mathcal{Y} = \mathcal{Z} \times_1 \mathbf{U}^{(1)} \times_2 \mathbf{U}^{(2)} \dots \times_P \mathbf{U}^{(P)}$, where $\mathcal{Y} \in \mathbb{R}^{J_1\times \dots \times J_P}$, and $\mathcal{Y}[j_1, \dots, j_P] = \sum_{i_1 = 1}^{I_1} \dots \sum_{i_P = 1}^{I_P} \mathcal{Z}[i_1, \dots, i_P] \prod_{p = 1}^{P} \mathbf{U}^{(p)}[j_p, i_p]$. 

\subsection{Related works}

\subsubsection{\bf Anomaly detection in dynamic graphs} 
\begin{figure}
\includegraphics[width=0.48\textwidth]{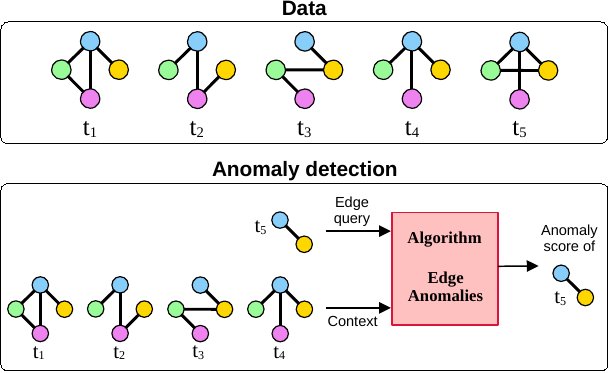}
\caption{Illustration of the query-context framework for anomaly detection in dynamic graphs. Algorithms can be seen as black-boxes that assess the anomaly level of a query structure based on contextual data. In the example, the query consists of a time-stamped edge at $t_5$, while the context consists of past graphs.} \label{fig:related-query_context_framework}
\end{figure}

The recent work of \cite{bautista.2024.mad} introduced a query-context framework for categorizing anomaly detection algorithms in dynamic graphs. In this framework, algorithms are seen as black boxes that receive two inputs, a query and a context, and output an anomaly score measuring the extent to which the query cannot be explained by the given context. The differences between algorithms thus essentially lie in: (i) the types of queries they assess; (ii) the contextual information they exploit; and (iii) the anomaly criterion they employ. Figure \ref{fig:related-query_context_framework} gives an illustration. Interestingly, this framework can be used to categorize the works in the literature according to the types of queries they assess, revealing that most approaches tackle queries representing structures defined at individual timestamps. 

For instance, although differing in the data they exploit and the anomaly criteria they employ, popular methods such as MIDAS \cite{bhatia.2022.midas}, F-FADE \cite{chang.2021.ffade}, TADDY \cite{liu.2023.taddy}, SEDANSPOT \cite{eswaran.2018.sedanspot}, AnomalyLLM \cite{liu.2024.anomalyllm} and COMEDY \cite{liu.2025.comedy} all aim to detect abnormal time-stamped edge queries. Similarly, methods like DSEDN \cite{bansal.2022.dsedn}, GEABS \cite{jiao.2021.geabs}, SLADE \cite{lee.2024.slade}, Flex-GAD \cite{chakraborty.2025.flex}, and STRIPE \cite{liu.2024.stripe} all tackle queries made of time-stamped nodes. Queries made of time-stamped graph snapshots have also been extensively considered. For instance, ANOMRANK \cite{yoon.2019.anomrank}, DYNANOM \cite{guo.2022.dynanom}, SPOTLIGHT \cite{eswaran.2018.spotlight}, LAD \cite{huang.2020.lad}, CADENCE \cite{mcneil.2023.cadence}, and Luo et al. \cite{luo.2025.graph} all aim to spot abnormal graph snapshots. We stress that these methods differ in the pathway taken towards a decision, but they share the query objective. Lastly, detection of abnormal time-stamped sub-graphs has been tackled by works like Tasn{\'a}di et al. \cite{tasnadi.2022.new}, TSAD \cite{jiang.2022.tsad} and MAD \cite{bautista.2024.mad}. However, as discussed in the Introduction, these algorithms, which flag structures at individual timestamps, are not well suited for the detection of window anomalies. 

A few algorithms based on tensor analysis have been proposed to cope with window queries. For instance, DTA \cite{sun.2006.dta} sees a temporal graph as a sequence of tensors. It uses historical tensors to train a joint Tucker factorization and flags an incoming query tensor as abnormal if it cannot be accurately reconstructed via the learned factors. TensorSplat \cite{koutra.2012.tensorsplat} applies the PARAFAC decomposition to the query in order to spot if the factor matrices diverge from historical ones. Leggas et al. \cite{leggas.2021.knowledge} see the temporal graph as a sequence of daily tensors. It applies a modified PARAFAC decomposition to historical data to extract expected factor behavior, flagging a window as abnormal if its factors diverge from such reference. WINTENDED \cite{fernandes.2021.wintended} assumes that abnormal windows are ones with densification events. It applies the CP-APR decomposition to query tensors and develops an algorithm to identify traces of subgraphs with high activity from the extracted factors. SWTF \cite{xu.2019.swtf} uses a tensor model that accounts for cyclic patterns within windows. It assumes that abnormal windows have most of their energy in the orthogonal complement of the subspace captured by a low-rank Tucker decomposition. However, these methods have two main limitations: (i) they omit the node attributes; and (ii) they rely on tensor factorizations that tend to perform poorly on highly sparse tensors. 

\subsubsection{\bf Dynamic graph representation learning} Anomaly detection algorithms often build upon dynamic graph representation learning methods. These methods aim to map graph structures into latent spaces that can be shaped to address the anomaly detection problem. However, it is not straightforward to build upon existing approaches to develop a window anomaly detector, as most methods focus on embedding individual time-stamped nodes, or graph snapshots rather than entire temporal windows. Namely, methods such as TGCN \cite{zhao.2019.tgcn}, JODIE \cite{kumar.2019.jodie}, EvolveGCN \cite{pareja.2020.evolvegcn}, TGAT \cite{xu.2020.tgat}, GraphMixer \cite{cong.2023.graphmixer}, DyGFormer \cite{yu.2023.dygformer}, and WinGNN \cite{zhu.2023.wingnn} focus on learning representations of time-stamped nodes. Other approaches like GraphERT \cite{beladev.2023.graphert}, tdGraphEmbed \cite{beladev.2020.tdgraphembed}, Wang et al. \cite{wang.2023.graph} and Melnyk et al. \cite{melnyk.2023.understanding} target the embedding of entire graph snapshots. A comprehensive survey of dynamic graph representation learning approaches can be found in \cite{feng.2026.survey}. 

In the literature, various tensor-based neural architectures have been proposed. These methods could potentially be adapted to embed entire temporal windows. However, adapting them for anomaly detection is not straightforward. For instance, RTGNN \cite{zhao.2023.rtgnn} is a tensor-based model capable of embedding multi-view graphs. However, it relies on a supervised training procedure tailored to classification. DSTGNN \cite{jia.2020.dstgnn} proposes a tensor network that uses graph convolutions to produce graph-informed representations of node signals within temporal windows. Yet, the graph is inferred from the features and the architecture is significantly oriented toward feature-level tasks. MA-NTAE \cite{zheng.2025.mantae} and DATC \cite{xie.2023.datc} are tensor-based autoencoders designed to learn compact latent representations through non-linear projections. However, neither method supports a joint encoding of the topology and the attributes. A recent survey of neural tensor decompositions can be found in \cite{zhao.2026.deep}.

\subsubsection{\bf Problem statement} The objective of this work is to develop an anomaly detection algorithm capable of assigning anomaly scores to queries of the form $\mathcal{G}_{\mathbb{W}}$, reflecting unusual behavior in the graph topology, node attributes, or both. To this end, we aim to design an autoencoder that maps inputs of the form $\mathcal{G}_{\mathbb{W}}$ into a compact latent representation $\mathcal{H}$, from which a reconstruction $\widehat{\mathcal{G}}_{\mathbb{W}}$ can be derived. After training on normal instances, anomaly scores can be obtained by measuring discrepancies between $\mathcal{G}_{\mathbb{W}}$ and $\widehat{\mathcal{G}}_{\mathbb{W}}$.

%% file: Sources/Proposed_method.tex
\section{Proposed Method}

\begin{figure*}
\includegraphics[width=\textwidth]{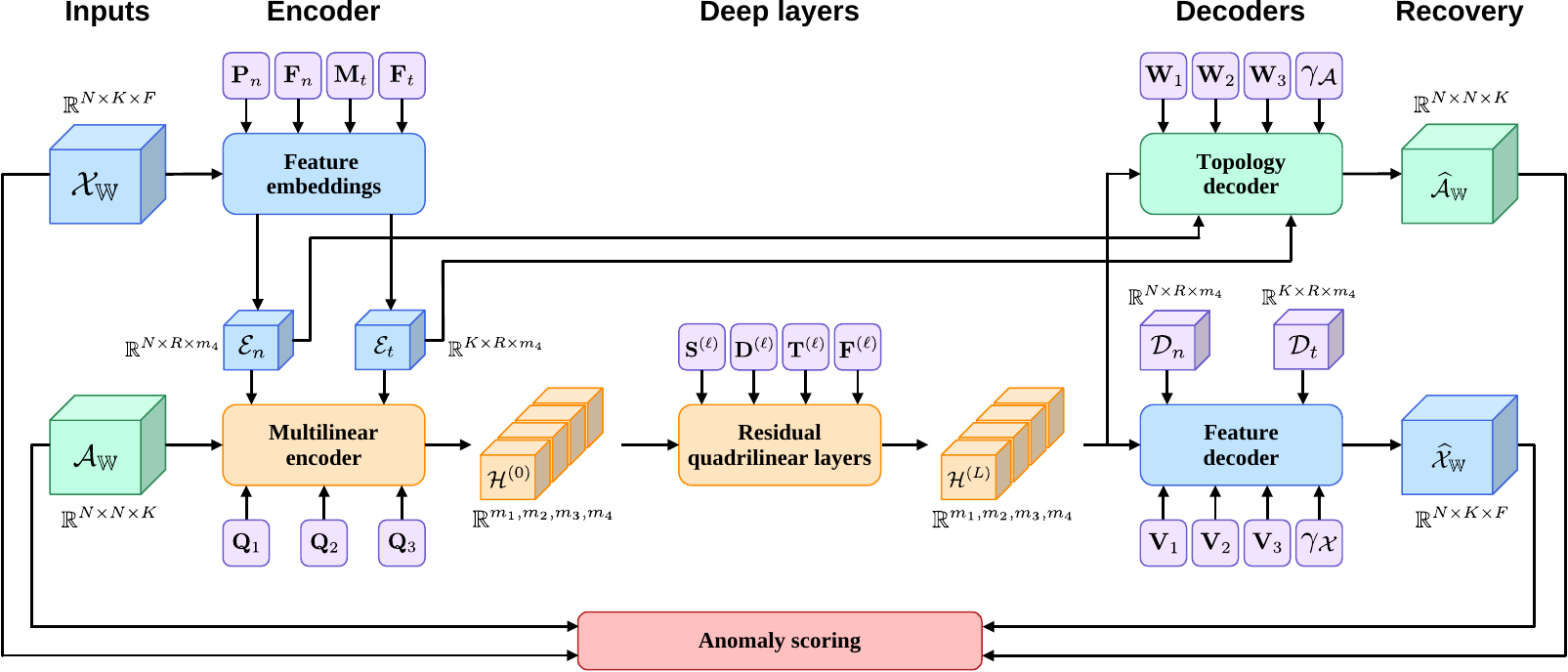}
    \caption{Architecture of MuTGN. The encoder constructs a latent core tensor through feature-conditioned projections of the adjacency tensor, followed by residual quadrilinear layers. Two decoder heads reconstruct the topology and node attributes, while reconstruction discrepancies are used for anomaly scoring. Colors distinguish feature-related components (blue), topology-related ones (green), latent representations (orange), learnable matrices and tensors (purple), and anomaly scoring (red). Biases and auxiliary parameters are omitted for clarity.} \label{fig:proposed-MuTGN_architecture}
\end{figure*}

In this section, we present MuTGN, a novel autoencoder architecture for window anomaly detection in attributed temporal graphs. MuTGN consists of four main components: (i) an encoder based on feature-conditioned projections of the adjacency tensor; (ii) deep layers based on residual quadrilinear operations; (iii) topology and feature decoder heads based on multilinear projections; and (iv) an anomaly scorer assessing discrepancies between observed and reconstructed data. The overall architecture of MuTGN is illustrated in Figure \ref{fig:proposed-MuTGN_architecture}. The following subsections detail its components and analyse its computational complexity.

\subsection{Encoder}

The goal of the encoder is to map each window $\mathcal{G}_{\mathbb{W}}$ into a compact latent representation. Since we assume that all windows come from the same temporal graph and that window abnormality is independent of their absolute position in time (see Figure \ref{fig:introduction-window_anomaly}), then we can address the problem by encoding their adjacency and feature tensors, $\mathcal{A}_{\mathbb{W}}$ and $\mathcal{X}_{\mathbb{W}}$. In the literature, a standard approach to encode $\mathcal{A}_{\mathbb{W}}$ is to project it into a core tensor $\mathcal{H} \in \mathbb{R}^{m_1 \times m_2 \times m_3}$ via the multilinear product $\mathcal{H} = \mathcal{A}_{\mathbb{W}} \times_1 \mathbf{U}_1 \times_2 \mathbf{U}_2 \times_3 \mathbf{U}_3$, where $\mathbf{U}_1 \in \mathbb{R}^{m_1 \times N}$, $\mathbf{U}_2 \in \mathbb{R}^{m_2 \times N}$, $\mathbf{U}_3 \in \mathbb{R}^{m_3 \times K}$ represent projection matrices. This encoding strategy is computationally appealing as the number of operations needed to compute the core is proportional to the number of nonzeros in $\mathcal{A}_{\mathbb{W}}$ and there is no need to materialize the zeros of $\mathcal{A}_{\mathbb{W}}$ in memory. Therefore, our aim is to build upon this formulation and to extend it to jointly encode $\mathcal{A}_{\mathbb{W}}$ and $\mathcal{X}_{\mathbb{W}}$ into a core $\mathcal{H}$. 

To this end, we propose (i) to add a fourth mode in $\mathcal{H}$ to encode feature information; and (ii) to condition the projection matrices on $\mathcal{X}_{\mathbb{W}}$. More precisely, since the first three modes of $\mathcal{H}$ represent the projection of information from source nodes, destination nodes and time, respectively, we propose to introduce a fourth mode that encodes the projection of feature information, yielding $\mathcal{H} \in \mathbb{R}^{m_1 \times m_2 \times m_3 \times m_4}$. Accordingly, we extend the projection matrices to tensors $\mathcal{U}_1 \in \mathbb{R}^{m_1 \times N \times m_4}$, $\mathcal{U}_2 \in \mathbb{R}^{m_2 \times N \times m_4}$, $\mathcal{U}_3 \in \mathbb{R}^{m_3 \times K \times m_4}$, so that the core is computed through slice-wise multilinear products: 
\begin{equation}\label{eq:proposed-4_mode_core}
        \mathcal{H}[a, b, c, d] = \sum_{u,v,k} \mathcal{A}_{\mathbb{W}}[u,v,k] \mathcal{U}_1[a, u, d] \mathcal{U}_2[b, v, d] \mathcal{U}_3[c, k, d],
\end{equation}
where $u,v \in \{0, \dots, N-1\}$ and $k \in \{0, \dots, K-1\}$. To make $\mathcal{H}$ aware of $\mathcal{X}_{\mathbb{W}}$, we propose to condition $\mathcal{U}_1$, $\mathcal{U}_2$, $\mathcal{U}_3$ on $\mathcal{X}_{\mathbb{W}}$. This can be naturally achieved by projecting $\mathcal{X}_{\mathbb{W}}$ into partial cores that represent $\mathcal{U}_1$, $\mathcal{U}_2$ and $\mathcal{U}_3$. That is,
\begin{equation}\label{eq:proposed-U_1}
    \mathcal{U}_1[a, u, d] = \sum_{k, f} \mathcal{X}_{\mathbb{W}}[u, k, f] \mathbf{T}_1[a,k] \mathbf{F}_1[d, f]
\end{equation}
\begin{equation}\label{eq:proposed-U_2}
    \mathcal{U}_2[b, v, d] = \sum_{k, f} \mathcal{X}_{\mathbb{W}}[v, k, f] \mathbf{T}_2[b,k] \mathbf{F}_2[d, f] 
\end{equation}
\begin{equation}\label{eq:proposed-U_3}
    \mathcal{U}_3[c, k, d] = \sum_{u, f} \mathcal{X}_{\mathbb{W}}[u, k, f] \mathbf{V}_3[c,u] \mathbf{F}_3[d, f]
\end{equation}
where $\mathbf{T}_1 \in \mathbb{R}^{m_1 \times K}$, $\mathbf{T}_2 \in \mathbb{R}^{m_2 \times K}$, $\mathbf{V}_3 \in \mathbb{R}^{m_3 \times N}$, $\mathbf{F}_i \in \mathbb{R}^{m4 \times F}$, for $i = 1, 2, 3$. 

If the projection matrices above are set as learnable, then Equations (\ref{eq:proposed-4_mode_core}), (\ref{eq:proposed-U_1}), (\ref{eq:proposed-U_2}), (\ref{eq:proposed-U_3}), already define a trainable encoder. However, such an encoder risks of overparameterization as the projection matrices involve a total of $K(m_1 + m_2) + N m_3 + 3 F m_4$ parametrs. This implies that increasing the core dimensions incurs a parameter cost proportional to $N$, $K$, or $F$, depending on the mode. While $F$ is often moderate, $N$ and $K$ can be very large in graphs with many nodes and fine-grained time resolution, which makes increases in core size very expensive in such regimes. To alleviate this issue, we propose to express the vertex and time contraction matrices as low-rank matrices that can be factorized as: $\mathbf{T}_1 = \mathbf{Q}_1 \mathbf{P}_1$, $\mathbf{T}_2 = \mathbf{Q}_2 \mathbf{P}_2$, $\mathbf{V}_3 = \mathbf{Q}_3 \mathbf{M}_3$, where $\mathbf{Q}_i \in \mathbb{R}^{m_i \times R}$, $\mathbf{P}_i \in \mathbb{R}^{R \times K}$, $\mathbf{M}_3 \in \mathbb{R}^{R \times N}$, with $R$ chosen sufficiently small. Substituting these factorizations in (\ref{eq:proposed-U_1}), (\ref{eq:proposed-U_2}), (\ref{eq:proposed-U_3}), results in:
\begin{equation}\label{eq:proposed-U_1_lowrank}
    \mathcal{U}_1[a, u, d] = \sum_{r} \left( \sum_{k, f} \mathcal{X}_{\mathbb{W}}[u, k, f] \mathbf{P}_1[r, k] \mathbf{F}_1[d, f] \right) \mathbf{Q}_1[a, r] 
\end{equation}
\begin{equation}\label{eq:proposed-U_2_lowrank}
    \mathcal{U}_2[b, v, d] = \sum_{r} \left( \sum_{k, f} \mathcal{X}_{\mathbb{W}}[v, k, f] \mathbf{P}_2[r, k] \mathbf{F}_2[d, f] \right) \mathbf{Q}_2[b, r] 
\end{equation}
\begin{equation}\label{eq:proposed-U_3_lowrank}
    \mathcal{U}_3[c, k, d] = \sum_{r} \left( \sum_{u, f} \mathcal{X}_{\mathbb{W}}[u, k, f] \mathbf{M}_3[r, u] \mathbf{F}_3[d, f] \right) \mathbf{Q}_3[c, r] 
\end{equation}

Two important consequences arise. Firstly, the low-rank factorization reduces the parameter count to $R(m_1 + m_2 + m_3 + 2K + N) + 3 F m_4$. Thus, increasing the core dimensions associated to the node and time modes now involves a cost proportional to $R$ rather than $N$ or $K$. Secondly, the terms within the parenthesis can be interpreted as a projection of $\mathcal{X}_{\mathbb{W}}$ into node and time embeddings. For instance, the term within the parenthesis in Eq. (\ref{eq:proposed-U_1_lowrank}) can be interpreted as follows: it first projects, through $\mathbf{F}_1$, the $F$-dimensional feature fector of each (node, time) pair into a new $m_4$-dimensional latent vector associated to the same (node, time) pair; then, for each node $u$, it aggregates the latent vectors associated to node $u$ over time, using $\mathbf{P}_1$ as weights that tune the influence of timestamps. For a fixed $r$, this yields a single $m_4$-dimensional representation of node $u$. The term within the parenthesis in Eq. (\ref{eq:proposed-U_2_lowrank}) also performs a node embedding, while the term in Eq. (\ref{eq:proposed-U_3_lowrank}) performs a time embedding, since $\mathbf{M}_3$ performs the weighted aggregation step over nodes.

The above insights therefore allow us to establish the definitive architecture of our encoder. Firstly, we let $\mathcal{E}_n \in \mathbb{R}^{N \times R \times m_4}$ denote a node embedding and $\mathcal{E}_t \in \mathbb{R}^{K \times R \times m_4}$ denote a time embedding, defined as:
\begin{equation}
    \begin{aligned}
        \mathcal{E}_n[u, r, d] = \frac{1}{\Omega_u} \sum_{k, f} \mathcal{X}_{\mathbb{W}}[u, k, f] &\mathbf{P}_n[r, k] \mathbf{F}_n[d, f] \\ &+ \mathbf{u}_n[u] + \mathbf{r}_n[r] + \mathbf{d}_n[d]
    \end{aligned}
\end{equation}
\begin{equation}
    \begin{aligned}
        \mathcal{E}_t[k, r, d] = \frac{1}{\Omega_k} \sum_{u, f} \mathcal{X}_{\mathbb{W}}[u, k, f] &\mathbf{M}_t[r, u] \mathbf{F}_t[d, f] \\ &+ \mathbf{k}_t[k] + \mathbf{r}_t[r] + \mathbf{d}_t[d],
    \end{aligned}
\end{equation}
where $\mathbf{P}_n \in \mathbb{R}^{R \times K}$, $\mathbf{M}_t \in \mathbb{R}^{R \times N}$, $\mathbf{F}_n, \mathbf{F}_t \in \mathbb{R}^{m_4 \times F}$; $\Omega_u = \max(1, |\{t \in \mathbb{W}: (u, t) \in \mathbb{P}_{\mathbb{W}} \}|)$ and $\Omega_k = \max(1, |\{u \in \mathbb{V}: (u, t_0 + k) \in \mathbb{P}_{\mathbb{W}} \}|)$ are normalization terms that control dominance of nodes and times with more observations; and $\mathbf{u}_n \in \mathbb{R}^{N}$, $\mathbf{k}_t \in \mathbb{R}^{K}$, $\mathbf{r}_n, \mathbf{r}_t \in \mathbb{R}^{R}$, $\mathbf{d}_n, \mathbf{d}_t \in \mathbb{R}^{m_4}$ are mode-wise bias terms that control the degenerte case in which $\mathcal{X}_{\mathcal{W}} = 0$. Secondly, we define the projection tensors as:
\begin{equation}
    \mathcal{U}_1[a, u, d] = \sum_{r} \mathcal{E}_n[u, r, d] \mathbf{Q}_1[a, r]
\end{equation}
\begin{equation}
    \mathcal{U}_2[b, v, d] = \sum_{r} \mathcal{E}_n[v, r, d] \mathbf{Q}_2[b, r]
\end{equation}
\begin{equation}
    \mathcal{U}_3[c, k, d] = \sum_{r} \mathcal{E}_t[k, r, d] \mathbf{Q}_3[c, r],
\end{equation}
where the node embedding tensor is shared among $\mathcal{U}_1$ and $\mathcal{U}_2$. The matrices $\mathbf{Q}_1$, $\mathbf{Q}_2$, $\mathbf{Q}_3$ here act as filters: they amplify or attenuate the contribution of each of the $R$ embeddings of a node/time when computing $\mathcal{U}_1$, $\mathcal{U}_2$, $\mathcal{U}_3$. Lastly, we build upon Eq. (\ref{eq:proposed-4_mode_core}) to construct the core as:
\begin{equation}
    \mathcal{Z}[a, b, c, d] =  \sum_{u,v,k} \mathcal{A}_{\mathbb{W}}[u,v,k] \mathcal{U}_1[a, u, d] \mathcal{U}_2[b, v, d] \mathcal{U}_3[c, k, d] 
\end{equation}
\begin{equation}
    \mathcal{H}^{(0)} = \text{LeakyReLU}\left( \text{LayerNorm}\left( \mathcal{Z} \right) \right),
\end{equation}
where LayerNorm plays a double role: (i) it controls dominance of windows with more observations; and (ii) it produces a non-trivial $\mathcal{H}^{(0)}$ in the case where $\mathcal{A}_{\mathbb{W}} = 0$. It is worth noting that our encoder effectively exploits the sparsity of the input data, as $\mathcal{E}_n$, $\mathcal{E}_t$, and $\mathcal{Z}$ can all be computed incrementally by accumulating contributions from individual observed feature vectors and edges, without needing to materialize the zero entries of $\mathcal{A}_{\mathbb{W}}$ and $\mathcal{X}_{\mathbb{W}}$.

\subsection{Deep layers}

The goal of the deep layers is to further structure the latent core tensor. We hypothesize that $\mathcal H^{(0)}$ already provides a meaningful initial encoding. Thus, we propose that subsequent layers progressively refine this core while preserving its four-mode organization. We achieve this through deep layers based on residual quadrilinear transformations:
\begin{equation}
    \mathcal{Z}^{(\ell)} = \text{LeakyReLU} \left( \mathcal{H}^{(\ell - 1)} \times_1 \mathbf{S}^{(\ell)} \times_2 \mathbf{D}^{(\ell)} \times_3 \mathbf{T}^{(\ell)} \times_4 \mathbf{F}^{(\ell)} + \mathcal{B}^{(\ell)} \right)
\end{equation}
\begin{equation}
    \mathcal{H}^{(\ell)} = \mathcal{H}^{(\ell - 1)} + \lambda^{(\ell)} \mathcal{Z}^{(\ell)}
\end{equation}
where $\ell = 1, \dots, L$ denotes the layer index, $\mathbf{S}^{(\ell)} \in \mathbb{R}^{m_1 \times m_1}$,  $\mathbf{D}^{(\ell)} \in \mathbb{R}^{m_2 \times m_2}$, $\mathbf{T}^{(\ell)} \in \mathbb{R}^{m_3 \times m_3}$, $\mathbf{F}^{(\ell)} \in \mathbb{R}^{m_4 \times m_4}$, are learnable mode-wise projection matrices, $\mathcal{B}^{(\ell)} \in \mathbb{R}^{m_1 \times m_2 \times m_3 \times m_4}$ is a learnable bias tensor, and $\lambda^{(\ell)} \in \mathbb{R}$ is a learnable scalar. 

In this formulation, $\mathcal Z^{(\ell)}$ represents a refinement extracted from $\mathcal H^{(\ell-1)}$ through the quadrilinear transformation, while $\lambda^{(\ell)}$ controls its contribution. We initialize $\lambda^{(\ell)}$ to zero, so that each layer initially behaves as an identity transformation, allowing the model to progressively learn the scope that each refinement layer must have. It is worth mentioning that alternative transformations could be used to compute the refinement $\mathcal Z^{(\ell)}$. A natural approach consists of flattening the core, applying a fully connected layer, and reshaping the result back into a tensor. However, such an approach would require $(\prod_{i=1}^{4}m_i)^2 + \prod_{i=1}^{4}m_i + 1$ learnable parameters, whereas our quadrilinear formulation requires only $\sum_{i=1}^{4}m_i^2 + \prod_{i=1}^{4}m_i + 1$, including the bias tensor and the residual coefficient.

\subsection{Decoders}

The goal of the decoders is to recover approximations of $\mathcal{A}_{\mathbb{W}}$ and $\mathcal{X}_{\mathbb{W}}$, respectively denoted $\widehat{\mathcal{A}}_{\mathbb{W}}$ and $\widehat{\mathcal{X}}_{\mathbb{W}}$, from $\mathcal{H}^{(L)}$. To this end, we propose two dedicated decoder heads, one recovering the topology and the other recovering the features. We detail them next.

\subsubsection{\bf Topology decoder} Our goal here is to recover $\widehat{\mathcal{A}}_{\mathbb{W}}$ from $\mathcal{H}^{(L)}$. Our decoding strategy builds on the observation that $\mathcal{H}^{(L)}$ contains $m_4$ latent encodings of the topology, one per slice $\mathcal H^{(L)}[:,:,:,d]$. Therefore, we propose to recover $\widehat{\mathcal{A}}_{\mathbb{W}}$ by first decoding the topological information carried by each slice and then by combining the resulting contributions through a learned weighted sum. We can achieve this through multilinear projections that mirror those used in the encoder: 
\begin{equation}\label{eq:proposed-latent_topology_decoding}
    \mathcal{Y}[u, v, k, d] = \sum_{a, b, c} \mathcal{H}^{(L)}[a, b, c, d] \mathcal{V}_1[a, u, d] \mathcal{V}_2[b, v, d] \mathcal{V}_3[c, k, d]
\end{equation}
\begin{equation}
    \widehat{\mathcal{A}}_{\mathbb{W}}[u, v, k] = \sum_{d} \gamma_{A}[d] \mathcal{Y}[u, v, k, d] + \mathbf{u}_A[u] + \mathbf{u}_A[v] + \mathbf{k}_A[k],
\end{equation}
where $\mathcal{V}_1 \in \mathbb{R}^{m_1 \times N \times m_4}$, $\mathcal{V}_2 \in \mathbb{R}^{m_2 \times N \times m_4}$, $\mathcal{V}_3 \in \mathbb{R}^{m_3 \times K \times m_4}$ are projection tensors, $\mathcal{Y} \in \mathbb{R}^{N \times N \times K \times m_4}$ contains slice-wise topology estimates, $\gamma_A \in \mathbb{R}^{m_4}$ is the learnable weight vector, and $\mathbf{u}_A \in \mathbb{R}^{N}$, $\mathbf{k}_A \in \mathbb{R}^{K}$ are mode-wise bias terms. 

As with the encoder, learning the projection tensors directly would involve a large number of parameters. To alleviate this issue, we adopt a similar low-rank factorization, expressing them as:
\begin{equation}
    \mathcal{V}_1[a, u, d] = \sum_{r} \mathcal{C}_n[u, r, d] \mathbf{W}_1[a, r]
\end{equation}
\begin{equation}
    \mathcal{V}_2[b, v, d] = \sum_{r} \mathcal{C}_n[v, r, d] \mathbf{W}_2[b, r]
\end{equation}
\begin{equation}
    \mathcal{V}_3[c, k, d] = \sum_{r} \mathcal{C}_t[k, r, d] \mathbf{W}_3[c, r]
\end{equation}
where $\mathcal{C}_n \in \mathbb{R}^{N \times R \times m_4}$, $\mathcal{C}_t \in \mathbb{R}^{K \times R \times m_4}$ are decoder-side node and time embeddings, and $\mathbf{W}_i \in \mathbb{R}^{m_i \times R}$ are learnable matrices. 

Since the encoder already constructs node and time embeddings, we propose to reuse them in the topology decoder by setting $\mathcal{C}_n = \mathcal{E}_n$ and $\mathcal{C}_t = \mathcal{E}_t$. This sharing allows the topology reconstruction objective to directly contribute to the learning of the embeddings, encouraging representations that are useful for both encoding and decoding. Therefore, our definitive topology decoder uses the projection tensors: 
\begin{equation}
    \mathcal{V}_1[a, u, d] = \sum_{r} \mathcal{E}_n[u, r, d] \mathbf{W}_1[a, r]
\end{equation}
\begin{equation}
    \mathcal{V}_2[b, v, d] = \sum_{r} \mathcal{E}_n[v, r, d] \mathbf{W}_2[b, r]
\end{equation}
\begin{equation}
    \mathcal{V}_3[c, k, d] = \sum_{r} \mathcal{E}_t[k, r, d] \mathbf{W}_3[c, r]
\end{equation}

The difference between the encoder projection tensors $\mathcal{U}_i$ and the topology decoder projection tensors $\mathcal{V}_i$ therefore stems from the filtering matrices $\mathbf{Q}_i$ and $\mathbf {W}_i$, which are learned independently to control how the shared embeddings contribute to encoding and reconstruction, respectively. Lastly, we stress that, for any query $(u,v,k)$, the value of $\widehat{\mathcal{A}}_{\mathbb{W}}[u,v,k]$ can be computed individually from the latent core and the corresponding node and time embeddings, without requiring the reconstruction of any other entry of $\widehat{\mathcal{A}}_{\mathbb{W}}$.

\subsubsection{\bf Features decoder} Our feature decoder builds on the observation that $\mathcal{Y}$, in Eq. (\ref{eq:proposed-latent_topology_decoding}), provides an $m_4$-dimensional latent representation of each triplet $(u, v, k)$. Yet, feature reconstruction calls for latent representations of node-time pairs $(u, k)$. We therefore revamp the multilinear formulation in (\ref{eq:proposed-latent_topology_decoding}) so that mode 1 and mode 2 in $\mathcal{H}^{(L)}$, which encode for source and destination nodes, project onto the same node $u$, as follows:
\begin{equation}
    \mathcal{F}[u, k, d] = \sum_{a,b,c} \mathcal{H}^{(L)}[a,b,c,d] \mathcal{P}_1[a,u,d] \mathcal{P}_2[b, u, d] \mathcal{P}_3[c,k,d],
\end{equation}
where $\mathcal{F} \in \mathbb{R}^{N \times K \times m_4}$, $\mathcal{P}_1 \in \mathbb{R}^{m_1 \times N \times m_4}$, $\mathcal{P}_2 \in \mathbb{R}^{m_2 \times N \times m_4}$, $\mathcal{P}_3 \in \mathbb{R}^{m_3 \times K \times m_4}$. Then, we project back to the original $F$ dimension through a learnable matrix $\gamma_X \in \mathbb{R}^{m_4 \times F}$ as:
\begin{equation}
    \widehat{\mathcal{X}}_{\mathbb{W}}[u, k, f] = \sum_{d} \gamma_X[d, f] \mathcal{F}[u, k, d] + \mathbf{u}_X[u] + \mathbf{k}_X[k] + \mathbf{f}_X[f],
\end{equation}
where $\mathbf{u}_X \in \mathbb{R}^{N}$, $\mathbf{k}_X \in \mathbb{R}^{K}$, $\mathbf{f}_X \in \mathbb{R}^{F}$ are mode-wise biases. As in our previous developments, we assume a low-rank modeling of the projection tensors, which we express as:
\begin{equation}
    \mathcal{P}_1[a, u, d] = \sum_{r} \mathcal{D}_n[u, r, d] \mathbf{V}_1[a, r]
\end{equation}
\begin{equation}
    \mathcal{P}_2[b, v, d] = \sum_{r} \mathcal{D}_n[v, r, d] \mathbf{V}_2[b, r]
\end{equation}
\begin{equation}
    \mathcal{P}_3[c, k, d] = \sum_{r} \mathcal{D}_t[k, r, d] \mathbf{V}_3[c, r]
\end{equation}
where $\mathcal{D}_n \in \mathbb{R}^{N \times R \times m_4}$, $\mathcal{D}_t \in \mathbb{R}^{K \times R \times m_4}$ are learnable node and time embeddings, and $\mathbf{V}_i \in \mathbb{R}^{m_i, R}$ are learnable matrices. 

Contrary to the topology decoder, we do not share the embeddings with the encoder, as this could create a reconstruction shortcut since $\mathcal{E}_n$ and $\mathcal{E}_t$ are extracted from $\mathcal{X}_{\mathbb W}$, the tensor we aim to reconstruct. Lastly, we highlight that feature vectors can be reconstructed individually for any node-time pair $(u,k)$, without requiring the reconstruction of the entire feature tensor $\widehat{\mathcal{X}}_{\mathbb{W}}$.

\subsection{Anomaly scoring and training}

\subsubsection{\bf Anomaly scoring}\label{sec:proposed-anomaly_scoring} The goal of the anomaly scorer is to assess discrepancies between the input and the reconstructed tensors. Since materializing the full reconstructed tensors in memory is prohibitively expensive, we measure discrepancies in carefully selected sets of entries. Namely, we define a topology evaluation set $\mathbb{S}_A = \mathbb{E}_{\mathbb{W}} \cup \mathbb{N}_{\mathbb{W}}$, where $\mathbb{N}_{\mathbb{W}}$ contains negative samples that are collected through three strategies: (i) triplets $(u, v, k) \notin \mathbb{E}_{\mathbb{W}}$ such that triplets of the form $(u, v, \cdot)$ appear in the window; (ii) triplets $(u, v, k) \notin \mathbb{E}_{\mathbb{W}}$ such that triplets of the form $(u, v, \cdot)$ have been observed in previous windows; and (iii) triplets $(u, v, k) \notin \mathbb{E}_{\mathbb{W}}$ so that $u$ and $v$ belong to the source and destination node populations, respectively. The size of $\mathbb{N}_{\mathbb{W}}$ is proportional to the size of $\mathbb{E}_{\mathbb{W}}$. Further details of our negative sampling strategies are provided in Appendix \ref{sec:appendix-negative_sampling}. In a nutshell, their aim is to identify absent edges that are structurally or temporally informative. Concerning  the features, we do not assume that unobserved node-time pairs necessarily correspond to zero-valued feature vectors (e.g., a bank account whose balance is not recorded at a given time does not necessarily have a zero balance). Thus, we define a feature evaluation set $\mathbb{S}_X = \mathbb{P}_{\mathbb{W}}$, containing only available feature observations. Given these evaluation sets, we define the topology and feature reconstruction losses as:

\begin{equation}
    \mathcal{L}_A(\mathcal{A}_{\mathbb{W}}, \widehat{\mathcal{A}}_{\mathbb{W}}) = \frac{\sum_{(u,v,k) \in \mathbb{S}_A} \left( \mathcal{A}_{\mathbb{W}}[u, v, k] - \widehat{\mathcal{A}}_{\mathbb{W}}[u, v, k] \right)^2}{ \max \{1, \sum_{(u,v,k) \in \mathbb{S}_A} \mathcal{A}_{\mathbb{W}}[u,v,k]^2 \} }
\end{equation}

\begin{equation}
    \mathcal{L}_X(\mathcal{X}_{\mathbb{W}}, \widehat{\mathcal{X}}_{\mathbb{W}}) = \frac{\sum_{(u,k) \in \mathbb{S}_X} \left\lVert \mathcal{X}_{\mathbb{W}}[u, k, :] - \widehat{\mathcal{X}}_{\mathbb{W}}[u, k, :] \right\rVert_2^2}{ \max \{1, \sum_{(u,k) \in \mathbb{S}_X} \left\lVert \mathcal{X}_{\mathbb{W}}[u,k,:] \right\rVert_2^2 \} }
\end{equation}
We further define a combined reconstruction loss that gives equal importance to each decoder head: 
\begin{equation}\label{eq:proposed-total_loss}
    \mathcal{L}(\mathcal{G}_{\mathbb{W}}, \widehat{\mathcal{G}}_{\mathbb{W}}) = \frac{\mathcal{L}_A(\mathcal{A}_{\mathbb{W}}, \widehat{\mathcal{A}}_{\mathbb{W}}) + \mathcal{L}_X(\mathcal{X}_{\mathbb{W}}, \widehat{\mathcal{X}}_{\mathbb{W}})}{2}
\end{equation}

While (\ref{eq:proposed-total_loss}) is useful for training and model selection, we detect anomalies by assessing whether the values of $\mathcal{L}_A$ and $\mathcal{L}_X$ are unusually large with respect to their historical values. Indeed, the topology and feature decoders may achieve different levels of reconstruction accuracy, which makes their raw losses difficult to compare. We thus assess abnormal reconstructions separately for each decoder, relative to its historical reconstruction performance. To this end, we maintain two fixed-lenght reference histories, $\operatorname{hist}{(\mathcal{L}_A)}$ and $\operatorname{hist}(\mathcal{L}_X)$, containing recent topology and feature reconstruction losses, respectively. These histories are initialized with the reconstruction losses measured on the validation dataset and subsequently updated as test windows are evaluated. Given these histories, we assess how unusually large each reconstruction loss is through modified Z-scores:

\begin{equation}
    z_j = \frac{ \mathcal L_j -\operatorname{median}(\operatorname{hist}(\mathcal{L}_j))}{1.4826 \operatorname{MAD}(\operatorname{hist}(\mathcal{L}_j))+\varepsilon}, \qquad j \in \{A, X\},
\end{equation}
where $\operatorname{MAD}$ denotes the median absolute deviation and $\varepsilon>0$ ensures numerical stability. Each reference history retains only the most recent losses and is updated only after the current window has been scored. Finally, we define the anomaly score of a window as:
\begin{equation}
    score(\mathcal{G}_{\mathbb{W}}) = \max \{ z_A, z_X \},
\end{equation}
allowing MuTGN to detect anomalies affecting either the topology or the node attributes, without requiring both reconstruction losses to be unusually large.

\subsubsection{\bf Training}\label{sec:proposed-training} We train MuTGN on windows reflecting normal behavior, using Eq. (\ref{eq:proposed-total_loss}) as the minimization objective. We randomly reserve a portion of the training windows for validation. For a fixed hyperparameter configuration, we optimize the model using Adam over shuffled mini-batches. At each epoch, we evaluate the average reconstruction loss on the validation dataset and apply early stopping when the validation loss ceases to improve, retaining the model weights that achieve the lowest validation loss. This model is then employed to initialize $\operatorname{hist}(\mathcal{L}_A)$ and $\operatorname{hist}(\mathcal{L}_X)$ using validation set losses. To select MuTGN's hyperparameters, we retain the configuration that achieves the lowest validation loss.

Temporal graphs are prone to concept drift, meaning that the weights learned during training may become progressively less representative of normal behavior. To account for such changes, we further employ an online fine-tuning mechanism that progressively adapts the model parameters as new windows are evaluated. In precise terms, at test time, a new incoming window is scored and then we perform additional Adam optimization steps on that window, maintaining (\ref{eq:proposed-total_loss}) as the minimization objective. This mechanism introduces two hyperparameters: the fine-tuning learning rate and the number of fine-tuning steps. We calibrate them on the validation dataset by processing validation windows sequentially and measuring each window's reconstruction loss before fine-tuning on that window. We progressively increase the learning rate and number of steps, retaining configurations that reduce the average validation loss until improvements saturate.

\subsection{Complexity analysis}

In this subsection, we demonstrate the computational efficiency of MuTGN. In particular, Proposition \ref{proposition:proposed-general_complexity} establishes its inference time complexity in terms of the input and architectural dimensions, while Corollary \ref{corollary:proposed-fixed_architecture_complexity} shows that, for fixed architectural dimensions, MuTGN scales linearly with the size of the window and the number of observations.

\begin{proposition}\label{proposition:proposed-general_complexity}
    Consider an input window comprising $N$ nodes, $K$ timestamps, $M_A$ observed timestamped edges, and $M_X$ observed node–time feature vectors, each of dimension $F$. Let $\mathbb{S}_A$ and $\mathbb{S}_X$ denote the topology and feature evaluation sets, respectively, satisfying $|\mathbb S_A | = (1+c) M_A$ and $|\mathbb S_X| = M_X$, where $c \geq 0$ is a fixed negative-sampling ratio. Moreover, assume that $\operatorname{hist}(\mathcal{L}_A)$ and $\operatorname{hist}(\mathcal{L}_X)$ have fixed length. A MuTGN model with latent core dimensions $(m_1, \dots, m_4)$, low-rank dimension $R$, and $L$ deep layers computes an anomaly score for the window in time:
    \begin{equation}\label{eq:proposed-general_complexity_bound}
    \begin{aligned}
        \mathcal{O}\Big( &M_X m_4 (F + R + R^3)  \\
        &+ M_A m_4 R^3  \\
        &+ (N + K) R m_4 \\
        &+ m_4(m_1 R^3 + m_1 m_2 R^2 + m_1 m_2 m_3 R ) \\
        &+ L m_1 m_2 m_3 m_4 (m_1 + m_2 + m_3 + m_4) \Big) 
    \end{aligned}
\end{equation}

\end{proposition}

\begin{corollary}\label{corollary:proposed-fixed_architecture_complexity}
    Under the notation and assumptions above, for fixed feature dimension and architectural hyperparameters, MuTGN produces an anomaly score in:
    \begin{equation}
        \mathcal{O}\left(N + K + M_A + M_X \right)
    \end{equation}
\end{corollary}

The proofs of Proposition \ref{proposition:proposed-general_complexity} and Corollary \ref{corollary:proposed-fixed_architecture_complexity} are deferred to Appendix \ref{sec:appendix-complexity}.

%% file: Sources/Numerical_experiments.tex
\section{Numerical Experiments}

This section evaluates MuTGN. Our experiments aim to address the following questions: {\bf Q1. Performance:} How accurately does MuTGN detect different types of window anomalies, and how does it compare with existing methods? {\bf Q2. Architecture:} How do MuTGN's architectural design choices influence its detection performance? {\bf Q3. Applicability:} Can MuTGN identify a documented Internet outage in a large-scale temporal graph derived from BGP data? 

\subsection{Experimental setup}

\subsubsection{\bf Datasets} We evaluate MuTGN on one synthetic and five real-world datasets. The synthetic dataset is generated using a variant of the modular activity-driven model \cite{han.2015.epidemic}, incorporating day-night activity cycles and three time-varying node features. The real-world datasets\footnote{Datasets available at \url{https://zenodo.org/records/7008205\#.Yv\_a\_3bMJPZ}.} are: (i) Wikipedia \cite{kumar.2019.predicting}, timestamped edits by users to Wikipedia pages, accompanied by 172-dimensional Linguistic Inquiry and Word Count (LIWC) feature vectors; (ii) Reddit \cite{kumar.2019.predicting}, timestamped user posts to subreddits, also accompanied by 172-dimensional LIWC vectors; (iii) MOOC \cite{kumar.2019.predicting}, timestamped student interactions with online course content, accompanied by 4-dimensional feature vectors; (iv) LastFM \cite{kumar.2019.predicting}, timestamped user listening records; and (v) Enron \cite{shetty.2004.enron}, email communications between employees. For Wikipedia, Reddit, and MOOC, features are originally associated with timestamped edges. We map them to node features by averaging the feature vectors of interactions incident to each node at each timestamp. We further standardize each multidimensional feature vector individually. LastFM and Enron do not contain features; we therefore use node degrees as a time-varying scalar features, supplementing positive-degree observations with sampled zero-degree node-time pairs at an 8:1 ratio. These are treated as valid zero-valued feature observations. The real-world datasets have a temporal resolution of one second. We partition the datasets into non-overlapping windows. For topology reconstruction, negative samples are drawn at a target 8:1 ratio using the three strategies described in Section \ref{sec:proposed-anomaly_scoring} in equal proportions. The windows are split chronologically into 70\% for training and 30\% for testing, with 15\% of the training portion reserved for validation. Table \ref{tab:experiments-mutgn_datasets} summarizes the resulting attributed temporal graphs and dataset splits. Appendix \ref{sec:appendix-synthetic} provides details about our synthetic model and its generation parameters. 

\begin{table}[t]
\centering
\caption{Datasets statistics.}
\label{tab:experiments-mutgn_datasets}
\setlength{\tabcolsep}{2pt}
\renewcommand{\arraystretch}{1.12}
\small
\begin{tabular}{@{}lrrrrrr@{}}
\toprule
 & Synthetic & Wikipedia & Reddit & MOOC & LastFM & Enron \\
\midrule
$N$ & 1,000 & 9,227 & 10,984 & 7,144 & 1,980 & 184 \\
$T$ & 900,000 & 2,678,374 & 2,678,391 & 2,572,087 & 137,107,268 & 113,740,400 \\
$F$ & 3 & 172 & 172 & 4 & 1 & 1 \\
$|\mathbb E|$ & 888,213 & 157,474 & 671,339 & 411,749 & 1,293,103 & 38,131 \\
$|\mathbb P|$ & 561,584 & 314,938 & 1,340,132 & 806,175 & 20,675,328 & 458,304 \\
$K$ & 600 & 600 & 600 & 600 & 86,400 & 86,400 \\
Train & 1,050 & 3,125 & 3,125 & 3,001 & 1,110 & 922 \\
Test & 450 & 1,339 & 1,339 & 1,286 & 476 & 395 \\
\bottomrule
\end{tabular}
\end{table}

\subsubsection{\bf Anomaly injection}
The selected datasets do not report window anomalies. We therefore investigate 11 types of injected window anomalies, grouped into three families:

\begin{itemize}[leftmargin=*]
    \item \textbf{Topology dynamics attacks:} (i) \emph{Time shuffling:} randomly permutes the timestamps of links between a subset of node pairs; (ii) \emph{Time shift:} circularly shifts the timestamps of links between a subset of node pairs; (iii) \emph{Link persistence:} makes randomly selected links persist over subsequent timestamps.

    \item \textbf{Topology structure attacks:} (iv) \emph{Uniform densification:} adds random links; (v) \emph{Structured densification:} adds links between previously observed node pairs; (vi) \emph{Sparsification:} removes randomly selected links; (vii) \emph{Rewiring:} redirects the destination of randomly selected links; (viii) \emph{T-ID swap:} exchanges the structural roles of randomly selected nodes.

    \item \textbf{Node feature attacks:} (ix) \emph{Feature shuffling:} randomly permutes the timestamps of features associated with a subset of nodes; (x) \emph{Feature shift:} circularly shifts the timestamps of features associated with a subset of nodes; and (xi) \emph{F-ID swap:} exchanges the features associated with randomly selected nodes.
\end{itemize}

For each dataset, we randomly select five test windows. Each attack scenario injects the same anomaly type into all five windows, with varying perturbation intensities. The selected windows remain the same across scenarios. Detailed attack definitions and parameters are provided in Appendix \ref{sec:appendix-attacks}.

\subsubsection{\bf Metrics} We evaluate performance using ROC AUC and PR AUC, considering attacked windows as positive samples. We report macro ROC AUC by averaging scores across attack types, datasets, or both, giving equal weight to each. Since the random PR AUC baseline depends on the proportion of anomalous test windows, we do not aggregate raw PR AUC values across datasets. Instead, we use the adjusted score
\begin{equation}
\mathrm{PR}_{\mathrm{adj}} = \frac{\mathrm{PR\ AUC}-\pi}{1-\pi},
\end{equation}
where $\pi$ denotes the fraction of anomalous test windows. This normalization maps the random baseline from $\pi$ to zero and a perfect score to one. We report raw PR AUC for individual datasets and adjusted PR AUC for cross-dataset aggregates.

\subsubsection{\bf Baselines} We compare MuTGN against six baseline methods: three window-level anomaly detectors, DTA \cite{sun.2006.dta}, SWTF \cite{xu.2019.swtf}, and WINTENDED \cite{fernandes.2021.wintended}, and three snapshot-level anomaly detectors, LAD \cite{huang.2020.lad}, MAD \cite{bautista.2024.mad}, and DynAnom \cite{guo.2022.dynanom}. For the latter, we aggregate the interactions within each window into a single graph snapshot. None of these methods handles node features; therefore, all baselines only use the adjaceny tensor and are evaluated only on topology anomalies. In all cases, we evaluate the produced anomaly scores before applying any hard decision. For each baseline and dataset, we explore a grid of hyperparameters and retain the configuration achieving the highest macro ROC AUC across the eight topology attack scenarios on the test set. MuTGN's model selection is performed using only normal validation data. We use the authors' implementations of LAD, MAD, and DynAnom, and our own implementations of DTA, SWTF, and WINTENDED. Hyperparameter grids and implementation details are provided in Appendix \ref{sec:appendix-model_selection}.

\subsection{Anomaly Detection Performance}

\begin{table*}[t]
\centering
\setlength{\tabcolsep}{4pt}
\renewcommand{\arraystretch}{1.10}
\caption{Detection performance by injected anomaly type, averaged across the six datasets. MuTGN reports mean $\pm$ standard deviation over five training seeds.}
\label{tab:experiments-attack_vs_method_summary}
\begin{tabular*}{\textwidth}{@{\extracolsep{\fill}}llrrrrrrr@{}}
\toprule
\textbf{Metric} & \textbf{Anomaly} & \textbf{MuTGN} & \textbf{DTA} & \textbf{WINTENDED} & \textbf{SWTF} & \textbf{LAD} & \textbf{MAD} & \textbf{DynAnom} \\
\midrule

\multirow{11}{*}{\shortstack[c]{\textbf{Macro}\\\textbf{ROC AUC}}}
    & Time shuffling & $\mathbf{0.852} \pm 0.011$ & $0.508$ & $0.557$ & $0.623$ & $0.618$ & $0.481$ & $0.492$ \\
    & Time shift & $\mathbf{0.887} \pm 0.004$ & $0.503$ & $0.523$ & $0.606$ & $0.610$ & $0.479$ & $0.492$ \\
    & Link persistence & $\mathbf{0.937} \pm 0.003$ & $0.511$ & $0.512$ & $0.411$ & $0.612$ & $0.481$ & $0.653$ \\
\cmidrule(lr){2-9}
    & Uniform densification & $\mathbf{0.889} \pm 0.003$ & $0.527$ & $0.524$ & $0.711$ & $0.579$ & $0.762$ & $0.757$ \\
    & Structured densification & $\mathbf{0.887} \pm 0.005$ & $0.516$ & $0.532$ & $0.681$ & $0.562$ & $0.478$ & $0.593$ \\
    & Sparsification & $\mathbf{0.654} \pm 0.010$ & $0.463$ & $0.490$ & $0.552$ & $0.463$ & $0.222$ & $0.290$ \\
    & Rewiring & $0.672 \pm 0.010$ & $0.568$ & $0.523$ & $0.612$ & $0.691$ & $\mathbf{0.700}$ & $0.508$ \\
    & T-ID swap & $\mathbf{0.699} \pm 0.016$ & $0.474$ & $0.523$ & $0.606$ & $0.605$ & $0.651$ & $0.628$ \\
\cmidrule(lr){2-9}
    & Feature shuffling & $\mathbf{0.784} \pm 0.020$ & \textemdash & \textemdash & \textemdash & \textemdash & \textemdash & \textemdash \\
    & Feature shift & $\mathbf{0.824} \pm 0.006$ & \textemdash & \textemdash & \textemdash & \textemdash & \textemdash & \textemdash \\
    & F-ID swap & $\mathbf{0.747} \pm 0.010$ & \textemdash & \textemdash & \textemdash & \textemdash & \textemdash & \textemdash \\
\midrule

\multirow{11}{*}{\shortstack[c]{\textbf{Macro}\\\textbf{adjusted}\\\textbf{PR AUC}}}
    & Time shuffling & $\mathbf{0.364} \pm 0.007$ & $0.001$ & $0.007$ & $0.009$ & $0.010$ & $0.036$ & $0.005$ \\
    & Time shift & $\mathbf{0.423} \pm 0.026$ & $0.001$ & $0.001$ & $0.013$ & $0.009$ & $0.036$ & $0.005$ \\
    & Link persistence & $\mathbf{0.571} \pm 0.029$ & $0.001$ & $0.000$ & $0.002$ & $0.017$ & $0.036$ & $0.221$ \\
\cmidrule(lr){2-9}
    & Uniform densification & $0.373 \pm 0.018$ & $0.002$ & $0.004$ & $0.047$ & $0.007$ & $\mathbf{0.396}$ & $0.279$ \\
    & Structured densification & $\mathbf{0.379} \pm 0.012$ & $0.001$ & $0.001$ & $0.012$ & $0.009$ & $0.036$ & $0.030$ \\
    & Sparsification & $\mathbf{0.094} \pm 0.012$ & $0.001$ & $0.003$ & $0.009$ & $0.006$ & $-0.001$ & $0.000$ \\
    & Rewiring & $0.124 \pm 0.013$ & $0.003$ & $0.001$ & $0.028$ & $0.014$ & $\mathbf{0.227}$ & $0.006$ \\
    & T-ID swap & $0.329 \pm 0.003$ & $0.001$ & $0.001$ & $0.006$ & $0.009$ & $0.266$ & $\mathbf{0.338}$ \\
\cmidrule(lr){2-9}
    & Feature shuffling & $\mathbf{0.345} \pm 0.011$ & \textemdash & \textemdash & \textemdash & \textemdash & \textemdash & \textemdash \\
    & Feature shift & $\mathbf{0.378} \pm 0.029$  & \textemdash & \textemdash & \textemdash & \textemdash & \textemdash & \textemdash \\
    & F-ID swap & $\mathbf{0.351} \pm 0.007$ & \textemdash & \textemdash & \textemdash & \textemdash & \textemdash & \textemdash \\
\bottomrule
\end{tabular*}
\end{table*}

This subsection addresses Q1. We evaluate MuTGN and the six baselines on the eight topology anomaly types across the six datasets. MuTGN is additionally evaluated on the three node-feature attacks. The results are summarized in Table \ref{tab:experiments-attack_vs_method_summary} and Table \ref{tab:experiments-dataset_vs_method_summary} from two complementary perspectives: Table \ref{tab:experiments-attack_vs_method_summary} reports macro ROC AUC and adjusted PR AUC across datasets for each anomaly type, while Table \ref{tab:experiments-dataset_vs_method_summary} reports ROC AUC and raw PR AUC averaged over the eight topology attacks for each dataset. 

Table \ref{tab:experiments-attack_vs_method_summary} shows that MuTGN detects the three topology dynamics attacks with high accuracy, significantly outperforming all baselines on both metrics. Concerning topology structure attacks, MuTGN also achieves strong results on both densification attacks, while remaining leader on sparsification and T-ID swaps in terms of ROC AUC. Rewiring is the only scenario in which MuTGN does not lead in any metric, yet its ROC AUC remains close to those of the leading method MAD and LAD. In general, sparsification, rewiring, and T-ID swap are challenging scenarios for all methods. MuTGN also shows good sensitivity to the three node-feature anomalies in both metrics. The window-based baselines perform poorly in general. Since all these methods build on low-rank representations of the adjacency tensor, we hypothesize that their performance reflects the difficulty of extracting informative low-rank patterns from highly sparse tensors. Snapshot-based methods, on the other hand, are by construction unable to detect time shuffling and time shift attacks, as their signatures are lost when the graph is aggregated. However, link persistence and topology structure attacks can still alter the aggregated graph, allowing these methods to remain sensitive to some of these anomalies.

Table \ref{tab:experiments-dataset_vs_method_summary} shows that MuTGN achieves the highest ROC AUC on five of the six datasets, consistently outperforming the baselines by large margins. Some baselines perform well on individual datasets, yet none achieves MuTGN's combination of high performance and consistency across these five datasets. Enron is the exception, where MuTGN performs poorly, although this dataset is difficult for all methods. Regarding PR AUC, MuTGN substantially outperforms the baselines on Synthetic and Reddit, more than doubling the second-best score. On Wikipedia and Enron, MuTGN is also a leader in terms of PR AUC. DynAnom achieves the hightest PR AUC on MOOC and LastFM, although MuTGN is a close second on MOOC and it still performs well relative to the random baseline of 0.01 on LastFM. 

Overall, these results show that MuTGN is an effective window anomaly detector across a broad range of anomalies, achieving particularly good performance on topology dynamics attacks, remaining very competitive on topology structure attacks, and reliably detecting node-feature anomalies. Moreover, it achieves consistently high ROC AUC across most datasets and PR AUC scores showing at least a ninefold improvement over the corresponding random baseline on all six datasets. Complete ROC AUC and PR AUC results for each dataset, attack type, and method are provided in Appendix \ref{sec:appendix-performance_tables}.

\begin{table*}[t]
\centering
\setlength{\tabcolsep}{4pt}
\renewcommand{\arraystretch}{1.13}
\caption{Detection performance by dataset, averaged over the eight topology attacks. MuTGN reports mean $\pm$ standard deviation over five training seeds.}
\label{tab:experiments-dataset_vs_method_summary}
\begin{tabular*}{\textwidth}{@{\extracolsep{\fill}}llrrrrrrr@{}}
\toprule
\textbf{Metric} & \textbf{Dataset} & \textbf{MuTGN} & \textbf{DTA} & \textbf{WINTENDED} & \textbf{SWTF} & \textbf{LAD} & \textbf{MAD} & \textbf{DynAnom} \\
\midrule
\multirow{6}{*}{\shortstack[c]{\textbf{Macro}\\\textbf{ROC AUC}}} 
    & Synthetic & $\mathbf{0.956} \pm 0.008$ & $0.633$ & $0.500$ & $0.594$ & $0.494$ & $0.723$ & $0.771$ \\
    & Wikipedia & $\mathbf{0.785} \pm 0.008$ & $0.543$ & $0.521$ & $0.638$ & $0.615$ & $0.473$ & $0.456$ \\
    & Reddit & $\mathbf{0.959} \pm 0.004$ & $0.458$ & $0.500$ & $0.689$ & $0.593$ & $0.713$ & $0.752$ \\
    & MOOC & $\mathbf{0.739} \pm 0.015$ & $0.426$ & $0.626$ & $0.365$ & $0.618$ & $0.270$ & $0.311$ \\
    & LastFM & $\mathbf{0.814} \pm 0.009$ & $0.503$ & $0.500$ & $0.654$ & $0.611$ & $0.584$ & $0.606$ \\
    & Enron & $0.605 \pm 0.015$ & $0.489$ & $0.491$ & $\mathbf{0.661}$ & $0.624$ & $0.429$ & $0.414$ \\
\midrule
\multirow{6}{*}{\shortstack[c]{\textbf{Macro}\\\textbf{PR AUC}}} 
    & Synthetic & $\mathbf{0.805} \pm 0.017$ & $0.019$ & $0.011$ & $0.033$ & $0.017$ & $0.336$ & $0.178$ \\
    & Wikipedia & $\mathbf{0.109} \pm 0.028$ & $0.004$ & $0.008$ & $0.046$ & $0.007$ & $0.082$ & $0.045$ \\
    & Reddit & $\mathbf{0.824} \pm 0.007$ & $0.004$ & $0.004$ & $0.008$ & $0.013$ & $0.180$ & $0.148$ \\
    & MOOC & $0.045 \pm 0.011$ & $0.004$ & $0.007$ & $0.005$ & $0.011$ & $0.004$ & $\mathbf{0.060}$ \\
    & LastFM & $0.120 \pm 0.017$ & $0.011$ & $0.011$ & $0.020$ & $0.016$ & $0.198$ & $\mathbf{0.261}$ \\
    & Enron & $\mathbf{0.121} \pm 0.025$ & $0.013$ & $0.018$ & $0.027$ & $0.041$ & $0.013$ & $0.012$ \\
\bottomrule
\end{tabular*}
\end{table*}

\subsection{Ablation Study}

We address Q2 through an ablation study investigating three architectural questions: (i) Does sharing node and time embeddings between the encoder and topology decoder improve detection performance compared with using independent embeddings? (ii) How important is normalizing these embeddings by the number of feature observations? (iii) What is the contribution of applying LayerNorm to the initial latent representation? To address these questions, we consider three architectural variants, each removing one of these components. For each dataset, all variants are trained from scratch using the same hyperparameter configuration as the reference model and evaluated on the 11 anomaly types. Figure \ref{fig:experiments-ablation_summary} reports macro ROC AUC and adjusted PR AUC aggregated across datasets and anomaly types, over five training seeds.

The results show that MuTGN's sharing of embeddings is crucial for anomaly detection performance. In particular, Figure \ref{fig:experiments-ablation_summary} shows that removing the sharing of node and time embeddings between the encoder and the topology decoder leads to a sharp drop in both ROC AUC and adjusted PR AUC. This suggests that it is beneficial for anomaly detection to make the topology decoder adaptive to the input features. Indeed, while the unshared model relies on fixed decoder projection tensors, MuTGN constructs these projection tensors based on the feature information of each input window. This adaptivity allows the decoder to base its topology reconstruction on the observed node attributes, introducing an inductive bias toward reconstructing topologies that are coherent with the observed features and hence increasing MuTGN's sensitivity to anomalies that disrupt this coherence. 

The normalization operations also contribute to detection performance. Removing the normalization of the embedding tensors $\mathcal{E}$ by the number of feature observations leads to a decrease in both metrics, particularly adjusted PR AUC. This suggests that, without normalization, some frequently observed nodes and times may induce disproportionately large embeddings, leading to less informative projection tensors. Removing LayerNorm also reduces detection performance, particularly in adjusted PR AUC. The architectural importance of LayerNorm goes beyond empirical benefits, as its learnable affine offset allows MuTGN to produce nonzero latent core tensors even when the input adjacency tensor is empty. Overall, these results support MuTGN's sharing and normalization scheme, with embedding sharing having the strongest impact on detection performance and embedding normalization being the most important normalization operation. Complete ROC AUC and PR AUC results for each dataset, attack type, and architectural variant are provided in Appendix \ref{sec:appendix-ablation_tables}.

\pgfplotsset{
  mutgn bars/.style={
    xbar,
    width=0.94\columnwidth,
    height=4cm,
    ymin=-0.5, ymax=3.5,
    y dir=reverse,
    ytick={0,1,2,3},
    yticklabels={
      MuTGN,
      No $\mathcal{E}$ norm.,
      No LayerNorm,
      No $\mathcal{E}$ sharing
    },
    yticklabel style={font=\small},
    xticklabel style={font=\small},
    xlabel style={font=\small},
    xmajorgrids,
    grid style={black!12},
    bar width=10pt,
    bar shift=0pt,
    clip=false
  },
  mutgn error bars/.style={
    error bars/x dir=both,
    error bars/x explicit,
    error bars/error bar style={
      draw=sharingblue,
      line width=0.6pt
    },
    error bars/error mark=|,
    error bars/error mark options={
      draw=sharingblue,
      line width=0.6pt,
      mark size=4pt
    }
  },
  mutgn reference errors/.style={
    mutgn error bars,
    error bars/error bar style={
      draw=mutgnblue,
      line width=0.6pt
    },
    error bars/error mark options={
      draw=mutgnblue,
      line width=0.6pt,
      mark size=4pt
    }
  }
}

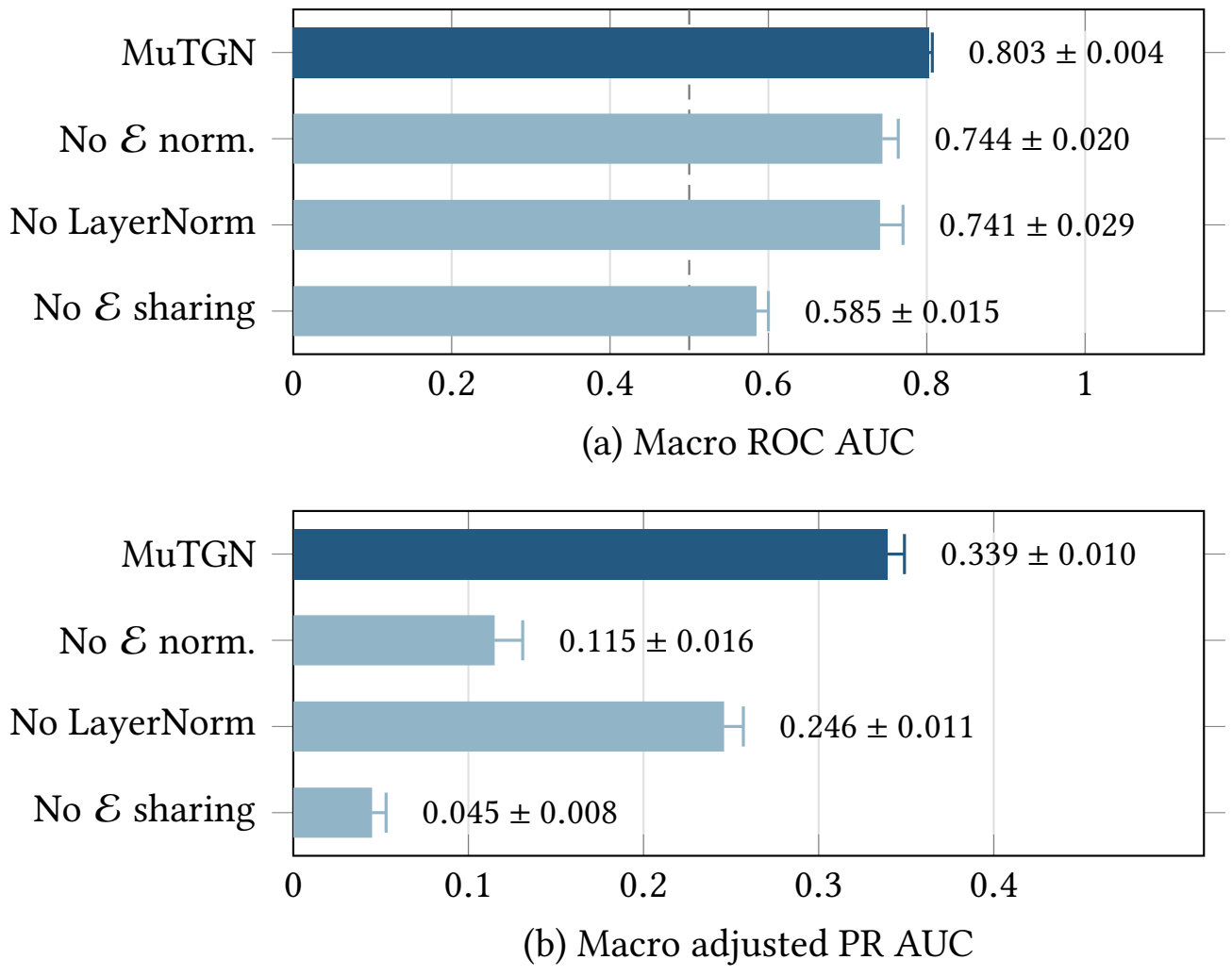
\begin{figure}[t]
\centering


\begin{tikzpicture}
\begin{axis}[
  mutgn bars,
  xmin=0, xmax=1.15,
  xtick={0,0.2,0.4,0.6,0.8,1},
  xlabel={(a) Macro ROC AUC}
]

  \draw[densely dashed,black!55]
    (axis cs:0.5,-0.5) --
    (axis cs:0.5,3.5);

  \addplot[
    draw=none,
    fill=sharingblue,
    mutgn error bars
  ] coordinates {
    (0.744,1) +- (0.020,0)
    (0.741,2) +- (0.029,0)
    (0.585,3) +- (0.015,0)
  };

  \addplot[
    draw=none,
    fill=mutgnblue,
    mutgn reference errors
  ] coordinates {
    (0.803,0) +- (0.004,0)
  };

  \node[
    anchor=west,
    font=\footnotesize,
    xshift=4pt
  ] at (axis cs:0.807,0)
    {$0.803 \pm 0.004$};

  \node[
    anchor=west,
    font=\footnotesize,
    xshift=4pt
  ] at (axis cs:0.764,1)
    {$0.744 \pm 0.020$};

  \node[
    anchor=west,
    font=\footnotesize,
    xshift=4pt
  ] at (axis cs:0.770,2)
    {$0.741 \pm 0.029$};

  \node[
    anchor=west,
    font=\footnotesize,
    xshift=4pt
  ] at (axis cs:0.600,3)
    {$0.585 \pm 0.015$};

\end{axis}
\end{tikzpicture}

\vspace{0.6em}


\begin{tikzpicture}
\begin{axis}[
  mutgn bars,
  xmin=0, xmax=0.52,
  xtick={0,0.1,0.2,0.3,0.4},
  xlabel={(b) Macro adjusted PR AUC}
]

  \addplot[
    draw=none,
    fill=sharingblue,
    mutgn error bars
  ] coordinates {
    (0.115,1) +- (0.016,0)
    (0.246,2) +- (0.011,0)
    (0.045,3) +- (0.008,0)
  };

  \addplot[
    draw=none,
    fill=mutgnblue,
    mutgn reference errors
  ] coordinates {
    (0.339,0) +- (0.010,0)
  };

  \node[
    anchor=west,
    font=\footnotesize,
    xshift=4pt
  ] at (axis cs:0.349,0)
    {$0.339 \pm 0.010$};

  \node[
    anchor=west,
    font=\footnotesize,
    xshift=4pt
  ] at (axis cs:0.131,1)
    {$0.115 \pm 0.016$};

  \node[
    anchor=west,
    font=\footnotesize,
    xshift=4pt
  ] at (axis cs:0.257,2)
    {$0.246 \pm 0.011$};

  \node[
    anchor=west,
    font=\footnotesize,
    xshift=4pt
  ] at (axis cs:0.053,3)
    {$0.045 \pm 0.008$};

\end{axis}
\end{tikzpicture}

\caption{Effect of architectural ablations on window anomaly detection. Bars report mean macro scores averaged across datasets and anomaly types. Error bars indicate one standard deviation over five training seeds.}
\label{fig:experiments-ablation_summary}

\end{figure}

\begin{figure}[t]
    \centering
    \includegraphics[width=\columnwidth]{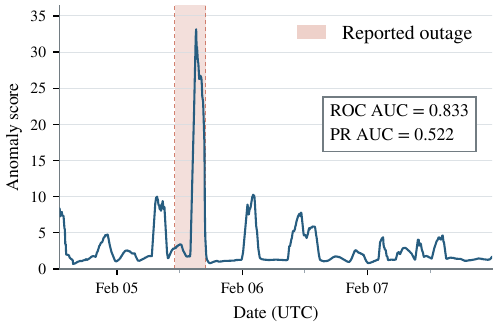}
    \caption{Anomaly scores produced by MuTGN on the AS-level Internet graph during a test period spanning from 13:00 on February 4 to 23:55 on February 7, 2023. An outage affecting an Italian Internet service provider was reported to have occurred on February 5, from 11:00 to 16:00. The shaded region identifies windows overlapping the documented outage period. MuTGN assigns its highest anomaly score to the window covering February 5 from 14:10 to 15:10, with scores remaining elevated through the end of the reported outage.}\label{fig:experiments-outage}
\end{figure}

\subsection{Case Study: Internet Outage Detection}

To address Q3, we evaluate MuTGN on a documented Internet outage. On February 5, 2023, an incident affecting Italy's largest Internet service provider disrupted connectivity for approximately 38\% of Italian Internet users, lasting roughly five hours from 11:00 to 16:00 \cite{justin.2026.madbgp}. We consider the BGP-derived temporal graph studied in \cite{justin.2026.madbgp}, covering February 3, 2023 at 00:00 to February 7, 2023 at 23:55. The graph contains 1,439 snapshots of the AS-level Internet topology at five-minute intervals, each containing approximately 80,000 nodes and 130,000 weighted links on average. This amounts to nearly 200 million time-stamped link observations, with highly heterogeneous weights ranging from $10^{-15}$ to $10^{-2}$. Details of the graph construction are provided in \cite{justin.2026.madbgp}. We use node degree as a time-varying scalar feature and do not sample negative links or zero-degree feature observations. We construct overlapping one-hour windows with a five-minute sliding step (one snapshot). To ensure that the outage period is excluded from training, we reserve the earliest 30\% of windows for training and the remaining windows for testing. A window is labeled anomalous if it overlaps the outage period. The resulting test set contains 996 windows, of which 71 are labeled anomalous.

Figure \ref{fig:experiments-outage} shows the time series of MuTGN's anomaly scores on the test windows, with the documented outage period highlighted in red. It can be seen that MuTGN is clearly sensitive to the Internet disruption, producing its highest anomaly scores during the outage. In particular, it assigns its highest score to the window covering February 5 from 14:10 to 15:10, and sustains high anomaly scores in subsequent windows until the end of the outage period. It can also be seen that time series produces noticeable score fluctuations outside the reported event period. These fluctuations may reflect baseline BGP dynamics, other undocumented events, or false positives. Nevertheless, MuTGN attributes the outage a peak score that is at least three times larger than these fluctuations, thus making it clearly distinguishable from such surrounding activity. Interestingly, the peak occurs with some delay relative to the reported outage onset. This behavior is not specific to MuTGN: a similar delayed peak is observed in the specialized BGP detector of \cite{justin.2026.madbgp} (see Figure~11 therein). As discussed in \cite{justin.2026.madbgp}, this delay may reflect a mismatch between the actual timing of BGP anomalies and the outage periods reported in the literature. 

Overall, this case study demonstrates that MuTGN can also be effective in realistic settings beyond injected anomalies, motivating further research into its applicability to real-world scenarios.

%% file: Sources/Appendix_complexity_analysis.tex
\section{Complexity Analysis}\label{sec:appendix-complexity}

\subsection{Proof of Proposition \ref{proposition:proposed-general_complexity}}

\begin{proof}

We tackle the proof by first establishing the complexity of the embedding and encoder, then by analyzing the complexity of the deep layers, and lastly the complexity of the decoders.

{\bf Embeddings and encoder.} In the embedding module, the feature vector projection from $F$ to $m_4$ dimensions has an asymptotic cost of $\mathcal{O}(M_x F m_4)$. The aggregation of these vectors involves a cost $\mathcal{O}(M_x R m_4)$ and the addition of bias terms along with the normalization step has a cost of $\mathcal{O}((N + K)Rm_4)$. Therefore, the cost of the node and time embeddings is:
    \begin{equation}\label{eq:appendix-prop_proof_c_embeddings}
    \operatorname{C_{embeddings}} = \mathcal{O} \Big( M_x m_4 (F + R) + (N + K)R m_4 \Big)
\end{equation}
    The encoder can be efficiently evaluated by first contracting the adjacency tensor via the embedding tensors, producing
    \begin{equation}
        \mathcal{\alpha}[a, b, c, d] = \sum_{(u, v, k) \in \mathbb{E}_{\mathbb{W}}} \mathcal{A}[u, v, k] \mathcal{E}_n[u, a, d] \mathcal{E}_n[v, b, d] \mathcal{E}_t[k, c, d]
    \end{equation}
    where $\mathcal{\alpha} \in \mathbb{R}^{R \times R \times R \times m_4}$ is an intermediate representation whose computation has cost $\mathcal{O}(M_A R^3 m_4)$. The mapping from $\mathcal{\alpha}$ to $\mathcal{H}^{(0)}$ involves the application of the filter matrices $\mathbf{Q}_1, \mathbf{Q}_2, \mathbf{Q}_3$ whose cost is $\mathcal{O}(m_4(m_1R^3 + m_1 m_2 R^2 + m_1 m_2 m_3R))$ if they are applied sequentially. Lastly, the LayerNorm and activation function involve a cost of $\mathcal{O}(m_1 m_2 m_3 m_4)$. Thus, the total cost of the encoder is:
    \begin{equation}\label{eq:appendix-prop_proof_c_encoder}
        \begin{aligned}
            \operatorname{C_{encoder}} = \mathcal{O}\Big( & M_A m_4 R^3 \\ 
            &+ m_4(m_1 R^3 + m_1 m_2 R^2 + m_1 m_2 m_3 R) \\ 
            &+ m_1 m_2 m_3 m_4  \Big)
        \end{aligned}
    \end{equation}

    {\bf Deep layers.} Each quadrilinear layer involves four mode-wise matrix products between the core tensor and projection matrices. If the projection matrices are applied succesively, the cost of the quadrilinear transform is $\mathcal{O}( \prod_{i = 1}^4 m_i \sum_{j = 1}^4 m_j$). The cost of the activation function, biases and residual connections are $\mathcal{O}(\prod_{i = 1}^4 m_i)$, which get dominated by the quadrilear operation. The total cost over $L$ deep layers is therefore:
    \begin{equation}\label{eq:appendix-prop_proof_c_layers}
    \operatorname{C_{layers}} = \mathcal{O} \Big( L m_1 m_2 m_3 m_4 (m_1 + m_2 + m_3 + m_4) \Big)
\end{equation}

    {\bf Decoders.} The output of the topology decoder can be efficiently computed by projecting $\mathcal{H}^{(L)}$ into an intermediate representation $\mathcal{\alpha}' \in \mathbb{R}^{R \times R \times R \times m_4}$ via the application of the filtering matrices $\mathbf{W}_1, \mathbf{W}_2, \mathbf{W}_3$. If the mode-wise products are applied succesively, the total cost of computing $\mathcal{\alpha}'$ is $\mathcal{O}(m_4(m_1 m_2 m_3 R + m_1 m_2 R^2 + m_1 R^3))$. For a given triplet $(u,v,k)$, the cost of computing $\mathcal{Y}[u, v, k, :]$ through slice-wise projections of $\mathcal{\alpha}'$ via $\mathcal{E}_n$ and $\mathcal{E}_t$ is $\mathcal{O}(m_4 R^3)$. Mapping $\mathcal{Y}[u,v,k,:]$ to $\widehat{\mathcal{A}}[u,v,k]$ costs $\mathcal{O}(m_4)$, which is dominated by the computation of $\mathcal{Y}[u,v,k,:]$. Under the assumption that $|\mathbb{S}_A| = (1 + c) M_A$, the cost of decoding all edges in the topology evaluation set is therefore:
    \begin{equation}\label{eq:appendix-prop_proof_c_adecoder}
        \operatorname{C_{A-decoder}} = \mathcal{O}\Big( M_A m_4 R^3 + m_4 (m_1 m_2 m_3 R + m_1 m_2 R^2 + m_1 R^3 ) \Big)
    \end{equation}
    The feature decoder can be computed following the same steps, using independent projection tensors and filtering matrices with the same dimensions as those of the topology decoder. The main difference is that, for a given node-time pair $(u,k)$, the feature decoder maps the latent $\mathcal{F}[u, k, :]$ to an $F$ dimensional vector. This operation cannot be established as dominated by the computation of $\mathcal{F}[u, k, :]$ and thus adds a cost of $\mathcal{O}(m_4 F)$. This implies that the cost of decoding all node-time paris in the feature evaluation set is:
    \begin{equation}\label{eq:appendix-prop_proof_c_xdecoder}
        \operatorname{C_{X-decoder}} = \mathcal{O}\Big( M_X m_4 (R^3 + F) + m_4 (m_1 m_2 m_3 R + m_1 m_2 R^2 + m_1 R^3 ) \Big)
    \end{equation}
Computing the reconstruction losses over the evaluation sets adds a cost of $\mathcal{O}(M_A + M_X F)$, which is dominated by the decoder costs. Moreover, since the reference histories have fixed lengths, computing the modified Z-scores and their maximum takes constant time. Therefore, the complexity bound follows by summing the costs in Eqs. (\ref{eq:appendix-prop_proof_c_embeddings}), (\ref{eq:appendix-prop_proof_c_encoder}), (\ref{eq:appendix-prop_proof_c_layers}), (\ref{eq:appendix-prop_proof_c_adecoder}), and (\ref{eq:appendix-prop_proof_c_xdecoder}), and absorbing constant factors.

\end{proof}

\subsection{Proof of Corollary \ref{corollary:proposed-fixed_architecture_complexity}}

\begin{proof} Note that when $F$, $R$, $L$ and the latent core dimensions ($m_1$, \dots, $m_4$) are fixed, all the factors involving these quantities in Eq. (\ref{eq:proposed-general_complexity_bound}) become constants. Thus, asymptotically, the complexity is
    \begin{equation}
        \mathcal{O}(N+K+M_A+M_X)
    \end{equation}
\end{proof}

%% file: Sources/Appendix_attack_descriptions.tex
\begin{table*}[!t]
\centering
\small
\setlength{\tabcolsep}{5pt}
\renewcommand{\arraystretch}{0.95}
\captionsetup{font=small,skip=3pt}
\caption{Attack intensities. The five perturbation fractions correspond to the five attacked windows, in chronological order. Fractions are expressed as percentages, while temporal shifts and persistence durations are expressed in timestamp units.}
\label{appendix:attack_intensities}
\begin{tabular*}{\textwidth}{@{\extracolsep{\fill}}llccc@{}}
\toprule
\textbf{Attack} &
\textbf{Dataset(s)} &
\textbf{Perturbation fractions (\%)} &
\textbf{Time shift} &
\textbf{Persistence duration} \\
\midrule
\multirow{3}{*}{\textbf{Time shuffling}}
& Synthetic                  & 10, 10, 10, 20, 20 & -- & -- \\
& Wikipedia, Reddit, MOOC    & 10, 20, 30, 40, 50 & -- & -- \\
& LastFM, Enron              & 10, 20, 30, 40, 50 & -- & -- \\
\midrule

\multirow{3}{*}{\textbf{Time shift}}
& Synthetic                  & 10, 10, 10, 20, 20 & 10   & -- \\
& Wikipedia, Reddit, MOOC    & 10, 20, 30, 40, 50 & 30   & -- \\
& LastFM, Enron              & 10, 20, 30, 40, 50 & 1800 & -- \\
\midrule

\multirow{3}{*}{\textbf{Link persistence}}
& Synthetic                  & 10, 10, 10, 20, 20 & -- & 5 \\
& Wikipedia, Reddit, MOOC    & 10, 20, 30, 40, 50 & -- & 5 \\
& LastFM, Enron              & 10, 20, 30, 40, 50 & -- & 5 \\
\midrule

\multirow{3}{*}{\textbf{Uniform densification}}
& Synthetic                  & 10, 10, 10, 20, 20 & -- & -- \\
& Wikipedia, Reddit, MOOC    & 10, 20, 30, 40, 50 & -- & -- \\
& LastFM, Enron              & 10, 20, 30, 40, 50 & -- & -- \\
\midrule

\multirow{3}{*}{\textbf{Structured densification}}
& Synthetic                  & 10, 10, 10, 20, 20 & -- & -- \\
& Wikipedia, Reddit, MOOC    & 10, 20, 30, 40, 50 & -- & -- \\
& LastFM, Enron              & 10, 20, 30, 40, 50 & -- & -- \\
\midrule

\multirow{3}{*}{\textbf{Sparsification}}
& Synthetic                  & 10, 10, 10, 20, 20 & -- & -- \\
& Wikipedia, Reddit, MOOC    & 10, 20, 30, 40, 50 & -- & -- \\
& LastFM, Enron              & 10, 20, 30, 40, 50 & -- & -- \\
\midrule

\multirow{3}{*}{\textbf{Rewiring}}
& Synthetic                  & 10, 10, 10, 20, 20 & -- & -- \\
& Wikipedia, Reddit, MOOC    & 10, 20, 30, 40, 50 & -- & -- \\
& LastFM, Enron              & 10, 20, 30, 40, 50 & -- & -- \\
\midrule

\multirow{3}{*}{\textbf{T-ID swap}}
& Synthetic                  & 10, 10, 10, 20, 20 & -- & -- \\
& Wikipedia, Reddit, MOOC    & 10, 20, 30, 40, 50 & -- & -- \\
& LastFM, Enron              & 10, 20, 30, 40, 50 & -- & -- \\
\midrule

\multirow{3}{*}{\textbf{Feature shuffling}}
& Synthetic                  & 10, 10, 10, 20, 20 & -- & -- \\
& Wikipedia, Reddit, MOOC    & 10, 20, 30, 40, 50 & -- & -- \\
& LastFM, Enron              & 10, 20, 30, 40, 50 & -- & -- \\
\midrule

\multirow{3}{*}{\textbf{Feature shift}}
& Synthetic                  & 10, 10, 10, 20, 20 & 10   & -- \\
& Wikipedia, Reddit, MOOC    & 10, 20, 30, 40, 50 & 30   & -- \\
& LastFM, Enron              & 10, 20, 30, 40, 50 & 1800 & -- \\
\midrule

\multirow{3}{*}{\textbf{F-ID swap}}
& Synthetic                  & 10, 10, 10, 20, 20 & -- & -- \\
& Wikipedia, Reddit, MOOC    & 10, 20, 30, 40, 50 & -- & -- \\
& LastFM, Enron              & 10, 20, 30, 40, 50 & -- & -- \\

\bottomrule
\end{tabular*}
\end{table*} 

\section{Attack Descriptions}\label{sec:appendix-attacks}

Detailed definitions of our attacks are given below. Their implementation is available in the code repository.

\begin{itemize}[leftmargin=*]

    \item \textbf{Time shuffling.}
    We identify distinct source-destination node pairs with recorded interactions within the attacked window and randomly select a fraction of such pairs. We then generate a random permutation of the $K$ window timestamps and apply it to every edge event associated with a selected pair. The same permutation is used for all selected pairs, while the source, destination, and weight of each event remain unchanged.

    \item \textbf{Time shift.}
    We randomly select a fraction of distinct source-destination node pairs with recorded interactions within the attacked window. The timestamps of all edge events associated with these pairs are circularly shifted by a fixed offset within the window. Events shifted beyond the end of the window wrap around to its beginning. The source, destination, and weight of each event remain unchanged.

    \item \textbf{Link persistence.}
    We randomly select a fraction of the edge events in the attacked window. For each selected event, we extend the edge's presence over subsequent timestamps, up to a specified duration. This is achieved by inserting additional occurrences of the same edge at consecutive timestamps, up to the persistence limit. The added edges preserve the weight of the selected edge. Existing occurrences are not duplicated, and no events are added beyond the end of the attacked window.

    \item \textbf{Uniform densification.}
We identify the sets of source and destination nodes based on activity up to and including the attacked window, and generate additional edges by independently sampling source and destination nodes uniformly from their respective sets. Each new edge is assigned a timestamp sampled uniformly within the window. The densification amount is specified as a fraction of the original number of edge events in the window. Collisions with existing edge events are avoided. New events are assigned the mean edge weight of the attacked window.

    \item \textbf{Structured densification.}
    We identify source-destination node pairs with recorded interactions within the attacked window, and generate additional edges by sampling uniformly from these pairs. Each new edge is assigned a timestamp sampled uniformly within the window. The densification amount is specified as a fraction of the original number of edge events in the window. Collisions with existing edge events are avoided. New events are assigned the mean edge weight of the attacked window.

    \item \textbf{Sparsification.}
    We randomly select a fraction of the edge events in the attacked window, uniformly and without replacement, and remove them from the edge list. The remaining events are left unchanged.

    \item \textbf{Rewiring.}
    We randomly select a fraction of the edge events in the attacked window. For each selected event, we replace its destination with a node sampled uniformly from the set of destination nodes observed in the dataset. The source, timestamp, and weight remain unchanged. Collisions with original edges or already rewired ones are avoided.

    \item \textbf{T-ID swap.}
    We randomly select a fraction of the nodes with activity in the attacked window and organize them into pairs. For each pair of nodes, their identifiers are exchanged in every edge event within the window, regardless of whether they appear as sources or destinations. The timestamps and weights of edges remain unchanged.

    \item \textbf{Feature shuffling.}
    We randomly select a fraction of the nodes having feature observations in the attacked window. We then generate a random permutation of the $K$ window timestamps and apply it to all feature observations associated with the selected nodes. The same permutation is used for all selected nodes. Node identifiers and feature vectors remain unchanged.

    \item \textbf{Feature shift.}
    We randomly select a fraction of the nodes having feature observations in the attacked window. The timestamps of their feature observations are circularly shifted by a fixed offset, wrapping around at the window boundaries. Node identifiers and feature vectors remain unchanged.

    \item \textbf{F-ID swap.}
    We randomly select a fraction of the nodes having feature observations in the attacked window and organize them into pairs. For each pair, we exchange their identifiers in all feature observations associated with either node within the window. Consequently, each node receives the feature observations originally associated with the other, while timestamps and feature vector values remain unchanged.
\end{itemize}

We recall that, for each dataset, we randomly select 5 test windows and attack them. The same attack is applied to all windows, with varying perturbation intensities. The selected windows do not vary accross attack scenarios. The perturbation intensities for each scenario and dataset are reported in Table~\ref{appendix:attack_intensities}.

%% file: Sources/Appendix_negative_sampling.tex
\section{Negative Sampling}\label{sec:appendix-negative_sampling}

Detailed descriptions of our negative sampling strategies are given below. Their implementation is available in the code repository. 

\begin{itemize}[leftmargin=*]

    \item \textbf{Current-window interactions.}
    We identify the set of distinct source-destination node pairs with recorded interactions within the current window. From this set, we sample pairs with probabilities proportional to the sum of their observed edge weights within the window. For a selected pair, we sample a timestamp uniformly among those at which no interaction is recorded between the two nodes. This mechanism produces negative samples for node pairs that interact within the current window, but not at the sampled timestamps. These negative samples complement positive ones by representing times at which active node pairs do not interact.

    \item \textbf{Historical interactions.}
    We maintain a bounded collection of distinct source-destination node pairs observed in preceding windows, retaining those with the largest maximum recorded edge weights. We sample pairs from this collection with probabilities proportional to their maximum observed edge weights. For a selected pair, we sample a timestamp with probability proportional to the historical edge activity at the corresponding timestamp position within the window. Collisions with existing edge events are avoided. These negative samples complement positive ones by representing missing edges between historically observed node pairs at historically active timestamp positions.

    \item \textbf{Source-destination population sampling.}
    We identify the sets of nodes appearing as sources and destinations in the dataset. We independently sample a source and a destination uniformly from their respective sets, and assign a timestamp sampled uniformly within the current window. Collisions with existing edge events are avoided. These negative samples complement positive ones by representing absent interactions between plausible source-destination node pairs, including pairs with no previously recorded interactions.

\end{itemize}

We conclude this section by noting that the three negative sampling mechanisms contribute equally to the number of negative samples. Moreover, we stress that negative sampling is performed after anomaly injection, using the resulting adjacency tensors. Consequently, edge events removed by sparsification become eligible for negative sampling, although their selection is not guaranteed.

%% file: Sources/Appendix_model_selection.tex
\begin{table*}[!t]
\centering
\small
\setlength{\tabcolsep}{2pt}
\renewcommand{\arraystretch}{1.15}
\captionsetup{font=small,skip=3pt}
\caption{Retained hyperparameter configurations for MuTGN and
the baselines on each dataset. \textit{Notation:}
MuTGN: $(R,L;m_1,m_2,m_3,m_4;\eta_{\mathrm{ft}},s_{\mathrm{ft}})$,
where $\eta_{\mathrm{ft}}$ and $s_{\mathrm{ft}}$ denote the
fine-tuning learning rate and number of steps;
DTA and WINTENDED: decomposition rank;
SWTF: $(F,R_1,R_2,R_3)$;
LAD: $(i_{\max},w_s,w_l)$;
MAD: maximum context length;
DynAnom: personalized PageRank parameter $\alpha$.}
\label{tab:model_selection}
\begin{tabular*}{\textwidth}{@{\extracolsep{\fill}}lccccccc@{}}
\toprule
\textbf{Dataset} &
\textbf{MuTGN} &
\textbf{DTA} &
\textbf{WINTENDED} &
\textbf{SWTF} &
\textbf{LAD} &
\textbf{MAD} &
\textbf{DynAnom} \\
\midrule
\textbf{Synthetic} &
$(8,2;16,16,16,2;0,0)$ &
64 & 2 & $(1,8,4,1)$ & $(2,10,40)$ & 30 & 0.15 \\
\textbf{Wikipedia} &
$(8,2;64,64,16,16;10^{-4},1)$ &
32 & 4 & $(2,32,32,1)$ & $(2,10,40)$ & 30 & 0.85 \\
\textbf{Reddit} &
$(8,2;16,16,16,64;10^{-6},1)$ &
32 & 2 & $(2,2,2,1)$ & $(16,5,20)$ & 30 & 0.85 \\
\textbf{MOOC} &
$(8,2;32,32,16,2;10^{-4},1)$ &
32 & 8 & $(4,2,2,2)$ & $(16,15,60)$ & 30 & 0.15 \\
\textbf{LastFM} &
$(8,3;8,8,64,1;10^{-5},5)$ &
8 & 2 & $(4,4,2,2)$ & $(4,5,20)$ & 30 & 0.15 \\
\textbf{Enron} &
$(8,2;8,8,64,1;10^{-4},5)$ &
8 & 2 & $(2,32,32,1)$ & $(16,5,20)$ & 30 & 0.15 \\
\bottomrule
\end{tabular*}
\end{table*}

\section{Model Selection}\label{sec:appendix-model_selection}

\subsection{Hyperparameter search.}
MuTGN is selected using the following procedure. Firstly, we explore the configuration space by progressively increasing model flexibility, using a fixed random seed of 42 for training. The two configurations achieving the lowest validation losses are shortlisted and subsequently retrained using five random seeds: 42, 123, 456, 789, and 2026. We select the configuration achieving the lowest mean validation loss across these runs. The fine-tuning learning rate and fine-tuning steps are then calibrated using the seed achieving the lowest validation loss, following the procedure described in Section \ref{sec:proposed-training}. For the baselines, we explore method-specific hyperparameter grids and retain the configuration maximizing the macro ROC AUC across the eight topology anomaly types for each dataset. Table \ref{tab:model_selection} reports the configurations retained for each method and dataset. The hyperparameters tuned for each method and their explored spaces are detailed below. 

\begin{itemize}[leftmargin=*]

\item {\bf MuTGN.} For each dataset, we explore the cartesian product between $R \in \{2, 4, 8\}$ and five configurations of $(L, m_1, m_2, m_3, m_4)$, given as follows: \emph{(i)} \textit{Synthetic:} (2, 4, 4, 4, 1), (2, 8, 8, 8, 1), (2, 8, 8, 8, 2), (2, 16, 16, 16, 2), and (3, 8, 8, 8, 2); \emph{(ii)} \textit{Wikipedia and Reddit:} (2, 8, 8, 8, 32), (2, 16, 16, 16, 32), (2, 16, 16, 16, 64), (2, 64, 64, 16, 16), and (3, 16, 16, 16, 64); \emph{(iii)} \textit{MOOC:} (2, 4, 4, 4, 2), (2, 8, 8, 8, 2), (2, 16, 16, 16, 2), (2, 32, 32, 16, 2), and (3,16,16,16,2); \emph{(iv)} \textit{LastFM and Enron:} (2, 4, 4, 4, 1), (2, 8, 8, 8, 1), (2, 16, 16, 16, 1), (2, 8, 8, 64, 1), and (3, 8, 8, 64, 1). For the Internet Outage we explored the cartiesan product between $R \in \{2, 4\}$ and (2, 4, 4, 4, 1), (2, 8, 8, 8, 1) for seeds 42 and 123 and reported the configuration with smallest validation loss, with no fine-tuning steps.

\item {\bf DTA.} We tune the dimensionality $R$ of the low-rank tensor approximation over the following values: \emph{(i)} \textit{Synthetic:} $R \in$ \{2, 4, 8, 16, 32, 64, 128, 256\}; \emph{(ii)} \textit{Wikipedia, Reddit, and MOOC:} $R \in$ \{8, 16, 32\}; \emph{(iii)} \textit{LastFM and Enron:} $R \in$ \{2, 4, 8\}.

\item {\bf WINTENDED.} We tune the dimensionality $R$ of the low-rank tensor approximation over the following values: \emph{(i)} \textit{Synthetic, Wikipedia, Reddit, and MOOC:} $R \in$ \{2, 4, 8, 16, 32\}; \emph{(ii)} \textit{LastFM and Enron:} $R \in$ \{2, 4, 8\}.

\item {\bf SWTF.} We tune the number of temporal folds $F$ and the decomposition ranks $(R_1,R_2,R_3)$. We explore the Cartesian product between $(F,R_3) \in$ \{(1, 1), (2, 1), (4, 1), (4, 2)\} and the following dataset-dependent values of $(R_1,R_2)$: \emph{(i)} \textit{Wikipedia, Reddit, MOOC, LastFM, and Enron:} \{(2, 2), (2, 4), (4, 2), (4, 4), (4, 8), (8, 4), (8, 8), (8, 16), (16, 8), (16, 16), (16, 32), (32, 2), (32, 16), (32, 32)\}; \emph{(ii)} \textit{Synthetic:} \{(2, 2), (2, 4), (4, 2), (4, 4), (4, 8), (8, 4), (8, 8), (8, 16), (16, 8), (16, 16), (16, 32), (32, 16), (32, 32), (32, 64), (64, 2), (64, 32), (64, 64)\}.

\item {\bf LAD.} We tune the number of retained singular values $i_{\max}$ and the lengths of the short-term and long-term historical windows, $w_s$ and $w_l$, respectively, measured in graph snapshots. We explore the Cartesian product between $i_{\max} \in$ \{2, 4, 8, 16\} and $(w_s,w_l) \in$ \{(5, 20), (10, 40), (15, 60), (20, 80)\} for all datasets.

\item {\bf MAD.} This method automatically selects the optimal context length given a user-defined upper bound. We set the maximum context length to 30 snapshots (representing aggregated windows) for all datasets.

\item {\bf DynAnom.} We tune the personalized PageRank parameter $\alpha$ over the following values: $\alpha \in$ \{0.15, 0.5, 0.85\} for all datasets.

\end{itemize}

\subsection{Implementation details}

We use the authors' implementations of LAD, MAD, and DynAnom, and our own implementations of DTA, WINTENDED, and SWTF. To obtain window-level anomaly scores, we made the following implementation choices and adaptations.

\begin{itemize}[leftmargin=*]

\item {\bf DTA.} We learn the projection matrices incrementally from normal training windows and keep them fixed during testing, using the same decomposition rank across the three tensor modes. We use the normalized squared reconstruction error as the anomaly score.

\item {\bf WINTENDED.} We follow the algorithm described in the original paper up to the computation of the three graph statistics for each candidate anomalous subgraph. The original method uses these statistics to verify anomaly candidates through binary decisions. Instead, we use them to produce a continuous anomaly score by summing the squared deviations of these statistics from their respective means, normalized by their variances over the remaining timestamps. We use the maximum score produced by a candidate as the anomaly score of the window.

\item {\bf SWTF.} We follow the algorithm described in the paper by reorganizing the window adjacency tensor into an edge-by-time-by-fold tensor. For each window, we fully estimate the projection matrices via HOSVD rather than updating previous estimates via Bi-SVD. While the original method associates mode-wise errors with different anomaly classes, we produce a single score by squaring the normalized reconstruction error of the joint reconstruction (see Eq. (5) in \cite{xu.2019.swtf}).

\item {\bf LAD.} We use the authors' code with the \texttt{difference=False} flag. To handle empty graphs, we added two missing safeguards: assigning zero singular values when the graph Laplacian is zero, and setting the cosine similarity to $1$ when either spectrum has negligible norm.

\item {\bf MAD.} We use the authors' graph anomaly detection branch of their code.

\item {\bf DynAnom.} We use the authors' graph anomaly detection branch of their code with the \texttt{track\_mode=all} flag. Among the proposed anomaly scores, we use $L_2$-based sum scores.

\end{itemize}

%% file: Sources/Appendix_synthetic_data.tex
\section{Synthetic Data Generation}\label{sec:appendix-synthetic}

Our generator is based on an extension of the modular activity-driven model of \cite{han.2015.epidemic}. The original model assigns a potential to each node for initiating interactions, with a community-based bias in the selected destination nodes. We extend it to incorporate periodic high-activity and low-activity cycles, heterogeneous community sizes, and time-varying node features. The specific configuration used for our experiments is given in Table \ref{appendix:synthetic_parameters}.

In particular, we generate $N$ nodes partitioned into communities of heterogeneous sizes. These sizes are drawn from a truncated discrete power-law distribution, $P(s)\propto s^{-\omega}$, with $s\in[s_{\min},s_{\max}]$, where any remaining group of nodes of size smaller than $s_{\min}$ is merged into the last community. Each node $u$ receives its own activity potential $a_u$, sampled from a truncated continuous power-law distribution, $p(a)\propto a^{-\gamma}$, with $a\in[a_{\min},a_{\max}]$. By letting $\ell$ denote a low-activity factor and $h$ a high-activity factor, at each timestamp $t$, a node $u$ becomes active with probability $p_u(t)=\min\{1,a_u c(t)\}$, where $c(t)=\ell+\frac{h-\ell}{2} \left[1+\sin\left(\frac{2\pi(t+\phi)}{P}\right)\right]$ introduces cyclic activity fluctuations with period $P$ and phase $\phi$. Once active, a node draws $m$ outgoing interactions from a Poisson distribution with mean $\lambda_m$. Each interaction is directed towards a node within the same community with probability $\mu$, or a node outside the community with probability $1-\mu$. The specific destination node is sampled with probability proportional to its associated activity potential within the selected group.

Each node is associated with a three-dimensional feature vector. The first feature consists of a periodic time-series with autoregressive noise of order one:
\begin{equation}
    x(u, t)[1] =q a_u\frac{1+\sin(2\pi(t+\phi_f)/P_f)}{2} +\epsilon_{u,t},
\end{equation}
where $P_f$ and $\phi_f$ control the feature's periodicity and phase, $q$ controls the signal-to-noise ratio, and the autoregressive noise is given as $\epsilon_{u,t}=0.7\epsilon_{u,t-1}+\eta_{u,t}$, where $\eta_{u,t}\sim\mathcal{N}(0,(a_u/q)^2)$. The second feature is a time-series that fluctuates around a community-shared baseline. It corresponds to a mean-reverting process:
\begin{equation}
x(u,t)[2] = (1-\kappa)x(u,t-1)[2] + \kappa b_u + \zeta_{u,t},
\end{equation}
where $b_u$ denotes the baseline value shared by nodes belonging to the same community. It is sampled independently for each community from $\mathcal{N}(0,C)$, with $C$ being the number of communities. The first term above therefore controls the smoothness of the temporal transition, while the second term enforces a return to the baseline value. The last term is distributed as $\zeta_{u,t}\sim\mathcal{N}(0,(a_u^2b_u^2/q^2))$ and introduces random fluctuations. The third feature is the instantaneous total degree:
\begin{equation}
x(u,t)[3] = d_t^{\mathrm{in}}(u)+d_t^{\mathrm{out}}(u).
\end{equation}
We sample the feature vectors at timestamps when a node has at least one outgoing interaction.

\begin{table}[!t]
\centering
\caption{Synthetic data generation parameters.}
\label{appendix:synthetic_parameters}
\begin{tabular}{ll}
\toprule
\textbf{Parameter} & \textbf{Value} \\
\midrule
Random seed & 42 \\
Number of nodes $N$ & 1000 \\
Number of timestamps $T$ & 900000 \\
Community size range $[s_{\min},s_{\max}]$ & $[20,50]$ \\
Community size exponent $\omega$ & 3 \\
Node activity range $[a_{\min},a_{\max}]$ & $[10^{-3},1]$ \\
Node activity exponent $\gamma$ & 3 \\
Intra-community probability $\mu$ & 0.85 \\
Mean active node interactions $\lambda_m$ & 1 \\
Activity period $P$ & 600 \\
Activity phase $\phi$ & 0 \\
Activity scaling range $[\ell,h]$ & $[10^{-3},1]$ \\
Signal-to-noise parameter $q$ & 8 \\
Mean-reversion factor $\kappa$ & 0.9 \\
Feature period $P_f$ & 600 \\
Feature phase $\phi_f$ & 0 \\
\bottomrule
\end{tabular}
\end{table}

%% file: Sources/Appendix_performance_tables.tex
\section{Detailed Performance Tables}\label{sec:appendix-performance_tables}

Tables \ref{appendix:performance_table_roc} and \ref{appendix:performance_table_pr} report the performance of MuTGN and the baselines across datasets and anomaly types, measured by ROC AUC and PR AUC, respectively.

\begin{table*}[!t]
\centering
\small
\setlength{\tabcolsep}{2.6pt}
\renewcommand{\arraystretch}{0.90}
\captionsetup{font=small,skip=3pt}
\caption{ROC AUC by dataset and anomaly type. MuTGN reports mean $\pm$ standard deviation over five training seeds.}
\label{appendix:performance_table_roc}
\begin{tabular*}{\textwidth}{@{\extracolsep{\fill}}llrrrrrrr@{}}
\toprule
\textbf{Dataset} & \textbf{Anomaly} & \textbf{MuTGN} & \textbf{DTA} & \textbf{WINTENDED} & \textbf{SWTF} & \textbf{LAD} & \textbf{MAD} & \textbf{DynAnom} \\
\midrule
\multirow{11}{*}{\textbf{Synthetic}} 
    & Time shuffling & $\mathbf{1.0000} \pm 0.0000 $ & $0.5411$ & $0.5000$ & $0.7524$ & $0.5276$ & $0.6665$ & $0.7209$ \\
    & Time shift & $\mathbf{1.0000} \pm 0.0000$ & $0.5685$ & $0.5000$ & $0.6049$ & $0.4939$ & $0.6697$ & $0.7209$ \\
    & Link persistence & $\mathbf{1.0000} \pm 0.0000 $ & $0.5996$ & $0.5000$ & $0.0004$ & $0.4467$ & $0.6674$ & $\mathbf{1.0000}$ \\
    & Uniform densification & $\mathbf{0.9993} \pm 0.0005$ & $0.7007$ & $0.5000$ & $0.8238$ & $0.4364$ & $0.9978$ & $0.9079$ \\
    & Structured densification & $\mathbf{0.9999} \pm 0.0002$ & $0.6539$ & $0.5000$ & $0.7793$ & $0.3182$ & $0.6683$ & $0.9537$ \\
    & Sparsification & $\mathbf{0.9012} \pm 0.0214$ & $0.6876$ & $0.5000$ & $0.6688$ & $0.5106$ & $0.2458$ & $0.4454$ \\
    & Rewiring & $0.7476 \pm 0.0459$ & $0.7263$  & $0.5000$ & $0.6036$ & $0.7649$ & $\mathbf{0.9164}$ & $0.7196$ \\
    & T-ID swap & $\mathbf{0.9999} \pm 0.0002$ & $0.5892$ & $0.5000$ & $0.5204$ & $0.4530$ & $0.9506$ & $0.6971$ \\
    & Feature shuffling & $\mathbf{1.0000} \pm 0.0000$ & \textemdash & \textemdash & \textemdash & \textemdash & \textemdash & \textemdash \\
    & Feature shift & $\mathbf{0.9998} \pm 0.0002$ & \textemdash & \textemdash & \textemdash & \textemdash & \textemdash & \textemdash \\
    & F-ID swap & $\mathbf{1.0000} \pm 0.0000 $ & \textemdash & \textemdash & \textemdash & \textemdash & \textemdash & \textemdash \\
\midrule
\multirow{11}{*}{\textbf{Wikipedia}} 
    & Time shuffling & $\mathbf{0.7324} \pm 0.0563$ & $0.5558$ & $0.5963$ & $0.6070$ & $0.6558$ & $0.3279$ & $0.3913$ \\
    & Time shift & $\mathbf{0.9023} \pm 0.0226$ & $0.5558$ & $0.4963$ & $0.6430$ & $0.6558$ & $0.3261$ & $0.3913$ \\
    & Link persistence & $\mathbf{0.9944} \pm 0.0047$ & $0.5558$ & $0.4963$ & $0.6316$ & $0.5499$ & $0.3268$ & $0.5186$ \\
    & Uniform densification & $0.8979 \pm 0.0225$ & $0.5558$ & $0.4963$ & $0.6725$ & $0.6558$ & $0.9273$ & $\mathbf{0.9694}$  \\
    & Structured densification & $\mathbf{0.8641} \pm 0.0296$ & $0.5558$ & $0.4963$ & $0.6561$ & $0.4936$ & $0.3295$ & $0.4090$ \\
    & Sparsification & $\mathbf{0.6226} \pm 0.0692$ & $0.5558$ & $0.5963$ & $0.5280$ & $0.5595$ & $0.1274$ & $0.1426$ \\
    & Rewiring & $0.5892 \pm 0.0480$ & $0.5558$ & $0.4963$ & $0.6946$ & $0.6948$ & $\mathbf{0.8396}$ & $0.4691$ \\
    & T-ID swap & $\mathbf{0.6772} \pm 0.0522$ & $0.4558$ & $0.4963$ & $0.6711$ & $0.6558$ & $0.5759$ & $0.3562$ \\
    & Feature shuffling & $\mathbf{0.7209} \pm 0.0414$ & \textemdash & \textemdash & \textemdash & \textemdash & \textemdash & \textemdash \\
    & Feature shift & $\mathbf{0.7146} \pm 0.0547$ & \textemdash & \textemdash & \textemdash & \textemdash & \textemdash & \textemdash \\
    & F-ID swap & $\mathbf{0.6000} \pm 0.0597$ & \textemdash & \textemdash & \textemdash & \textemdash & \textemdash & \textemdash \\
\midrule
    \multirow{11}{*}{\textbf{Reddit}} 
    & Time shuffling & $\mathbf{0.9999} \pm 0.0001$ & $0.4661$ & $0.5000$ & $0.6254$ & $0.5627$ & $0.6934$ & $0.7261$ \\
    & Time shift & $\mathbf{0.9999} \pm 0.0001$ & $0.4058$ & $0.5000$ & $0.6756$ & $0.5633$ & $0.6943$ & $0.7261$ \\
    & Link persistence & $\mathbf{1.0000} \pm 0.0000$ & $0.4285$ & $0.5000$ & $0.4639$ & $0.6840$ & $0.6945$ & $0.8468$ \\
    & Uniform densification & $\mathbf{1.0000} \pm 0.0000$ & $0.4226$ & $0.5000$ & $0.8240$ & $0.5897$ & $0.9121$ & $0.9402$\\
    & Structured densification & $\mathbf{1.0000} \pm 0.0000$ & $0.4018$ & $0.5000$ & $0.7892$ & $0.6214$ & $0.6943$ & $0.8343$ \\
    & Sparsification & $0.7035 \pm 0.0241$ & $0.4966$ & $0.5000$ & $\mathbf{0.7375}$ & $0.4343$ & $0.2852$ & $0.3766$ \\
    & Rewiring & $\mathbf{0.9699} \pm 0.0083$ & $0.6415$ & $0.5000$ & $0.6466$ & $0.7279$ & $0.8541$ & $0.7289$ \\
    & T-ID swap & $\mathbf{0.9999} \pm 0.0001$ & $0.3987$ & $0.5000$ & $0.7536$ & $0.5636$ & $0.8738$ & $0.8355$ \\
    & Feature shuffling & $\mathbf{0.9996} \pm 0.0003$ & \textemdash & \textemdash & \textemdash & \textemdash & \textemdash & \textemdash \\
    & Feature shift & $\mathbf{0.9998} \pm 0.0004$ & \textemdash & \textemdash & \textemdash & \textemdash & \textemdash & \textemdash \\
    & F-ID swap & $\mathbf{0.9999} \pm 0.0003$ & \textemdash & \textemdash & \textemdash & \textemdash & \textemdash & \textemdash \\
\midrule
\multirow{11}{*}{\textbf{MOOC}} 
    & Time shuffling & $\mathbf{0.7916} \pm 0.0122$ & $0.4794$ & $0.6891$ & $0.4226$ & $0.6899$ & $0.2710$ & $0.2637$ \\
    & Time shift & $\mathbf{0.7830} \pm 0.0124$ & $0.4794$ & $0.6891$ & $0.3852$ & $0.6899$  & $0.2698$ & $0.2637$ \\
    & Link persistence & $\mathbf{0.8416} \pm 0.0027$ & $0.4794$ & $0.6251$ & $0.2515$ & $0.6944$ & $0.2681$ & $0.3319$ \\
    & Uniform densification & $\mathbf{0.7996} \pm 0.0055$ & $0.4794$ & $0.5896$ & $0.4798$ & $0.6163$ & $0.3533$ & $0.3916$ \\
    & Structured densification & $\mathbf{0.8048} \pm 0.0061$ & $0.4794$ & $0.6473$ & $0.4550$ & $0.7173$ & $0.2689$ & $0.3138$ \\
    & Sparsification & $\mathbf{0.6855} \pm 0.0215$ & $0.1365$ & $0.3907$ & $0.2180$ & $0.1599$ & $0.1621$ & $0.1933$ \\
    & Rewiring & $\mathbf{0.7552} \pm 0.0742$ & $0.4794$ & $0.6891$ & $0.3432$ & $0.6870$ & $0.2851$ & $0.2645$ \\
    & T-ID swap & $0.4540 \pm 0.0778$ & $0.3970$ & $0.6891$ & $0.3668$ & $\mathbf{0.6899}$  & $0.2798$ & $0.4681$ \\
    & Feature shuffling & $\mathbf{0.7259} \pm 0.0776$ & \textemdash & \textemdash & \textemdash & \textemdash & \textemdash & \textemdash \\
    & Feature shift & $\mathbf{0.7211} \pm 0.0480$ & \textemdash & \textemdash & \textemdash & \textemdash & \textemdash & \textemdash \\
    & F-ID swap & $\mathbf{0.6585} \pm 0.0194$ & \textemdash & \textemdash & \textemdash & \textemdash & \textemdash & \textemdash \\
\midrule
\multirow{11}{*}{\textbf{LastFM}} 
    & Time shuffling & $\mathbf{0.9581} \pm 0.0069$ & $0.5032$ & $0.5000$ & $0.6484$ & $0.6212$ & $0.5261$ & $0.4611$ \\
    & Time shift & $\mathbf{0.9505} \pm 0.0059$ & $0.5032$ & $0.5000$ & $0.6696$ & $0.6212$ & $0.5197$ & $0.4611$ \\
    & Link persistence & $\mathbf{0.9930} \pm 0.0038$ & $0.5032$ & $0.5000$ & $0.4675$ & $0.6204$ & $0.5291$ & $0.7495$ \\
    & Uniform densification & $\mathbf{0.9165} \pm 0.0120$ & $0.5032$ & $0.5000$ & $0.7482$ & $0.6166$ & $0.8123$ & $0.8306$ \\
    & Structured densification & $\mathbf{0.9166} \pm 0.0091$ & $0.5032$ & $0.5000$ & $0.6811$ & $0.5699$ & $0.5261$ & $0.6246$ \\
    & Sparsification & $0.5923 \pm 0.0258$ & $0.5032$ & $0.5000$ & $\mathbf{0.6369}$ & $0.5397$ & $0.1902$ & $0.2603$ \\
    & Rewiring & $0.5758 \pm 0.0250$ & $0.5032$ & $0.5000$ & $0.7074$ & $0.6777$ & $\mathbf{0.7864}$ & $0.4620$ \\
    & T-ID swap & $0.6111 \pm 0.0285$ & $0.5032$ & $0.5000$ & $0.6747$ & $0.6212$ & $0.7822$ & $\mathbf{1.0000}$ \\
    & Feature shuffling & $\mathbf{0.8545} \pm 0.0058$ & \textemdash & \textemdash & \textemdash & \textemdash & \textemdash & \textemdash \\
    & Feature shift & $\mathbf{0.8910} \pm 0.0138$ & \textemdash & \textemdash & \textemdash & \textemdash & \textemdash & \textemdash \\
    & F-ID swap & $\mathbf{0.6996} \pm 0.0180$ & \textemdash & \textemdash & \textemdash & \textemdash & \textemdash & \textemdash \\
\midrule
\multirow{11}{*}{\textbf{Enron}} 
    & Time shuffling & $0.6285 \pm 0.0205$ & $0.5023$ & $0.5579$ & $\mathbf{0.6821}$ & $0.6533$ & $0.3982$ & $0.3905$ \\
    & Time shift & $\mathbf{0.6865} \pm 0.0138$ & $0.5023$ & $0.4526$ & $0.6605$ & $0.6379$ & $0.3951$ & $0.3905$ \\
    & Link persistence & $\mathbf{0.7945} \pm 0.0167$ & $0.5023$ & $0.4526$ & $0.6515$ & $0.6769$ & $0.3992$ & $0.4741$ \\
    & Uniform densification & $\mathbf{0.7233} \pm 0.0182$ & $0.5023$ & $0.5579$ & $0.7156$ & $0.5610$ & $0.5708$ & $0.5028$ \\
    & Structured densification & $\mathbf{0.7390} \pm 0.0119$ & $0.5023$ & $0.5472$ & $0.7272$ & $0.6523$ & $0.3836$ & $0.4197$ \\
    & Sparsification & $0.4203 \pm 0.0340$ & $0.3956$ & $0.4526$ & $0.5200$ & $\mathbf{0.5718}$ & $0.3241$ & $0.3231$ \\
    & Rewiring & $0.3932 \pm 0.0266$ & $0.5023$ & $0.4526$ & $\mathbf{0.6787}$ & $0.5949$ & $0.5182$ & $0.4038$ \\
    & T-ID swap & $0.4547 \pm 0.0568$ & $0.5023$ & $0.4526$ & $\mathbf{0.6490}$ & $0.6477$ & $0.4415$ & $0.4095$ \\
    & Feature shuffling & $\mathbf{0.4059} \pm 0.0486$ & \textemdash & \textemdash & \textemdash & \textemdash & \textemdash & \textemdash \\
    & Feature shift & $\mathbf{0.6184} \pm 0.0134$ & \textemdash & \textemdash & \textemdash & \textemdash & \textemdash & \textemdash \\
    & F-ID swap & $\mathbf{0.5227} \pm 0.0261$ & \textemdash & \textemdash & \textemdash & \textemdash & \textemdash & \textemdash \\
\bottomrule
\end{tabular*}
\end{table*}

\begin{table*}[!t]
\centering
\small
\setlength{\tabcolsep}{2.6pt}
\renewcommand{\arraystretch}{0.90}
\captionsetup{font=small,skip=3pt}
\caption{PR AUC by dataset and anomaly type. MuTGN reports mean $\pm$ standard deviation over five training seeds.}
\label{appendix:performance_table_pr}
\begin{tabular*}{\textwidth}{@{\extracolsep{\fill}}llrrrrrrr@{}}
\toprule
\textbf{Dataset} & \textbf{Anomaly} & \textbf{MuTGN} & \textbf{DTA} & \textbf{WINTENDED} & \textbf{SWTF} & \textbf{LAD} & \textbf{MAD} & \textbf{DynAnom} \\
\midrule
\multirow{11}{*}{\textbf{Synthetic}} 
    & Time shuffling & $\mathbf{1.0000} \pm 0.0000$ & $0.0150$ & $0.0111$ & $0.0323$ & $0.0225$ & $0.2186$ & $0.0377$ \\
    & Time shift & $\mathbf{1.0000} \pm 0.0000$ & $0.0152$ & $0.0111$ & $0.0623$ & $0.0136$ & $0.2192$ & $0.0377$ \\
    & Link persistence & $\mathbf{1.0000} \pm 0.0000$ & $0.0170$ & $0.0111$ & $0.0067$ & $0.0113$ & $0.2187$ & $\mathbf{1.0000}$ \\
    & Uniform densification & $\mathbf{0.9555} \pm 0.0290$ & $0.0209$ & $0.0111$ & $0.0407$ & $0.0121$ & $0.9000$ & $0.1040$ \\
    & Structured densification & $\mathbf{0.9933} \pm 0.0149$ & $0.0185$ & $0.0111$ & $0.0306$ & $0.0093$ & $0.2188$ & $0.1546$ \\
    & Sparsification & $\mathbf{0.4694} \pm 0.1211$ & $0.0201$ & $0.0111$ & $0.0349$ & $0.0206$ & $0.0105$ & $0.0144$ \\
    & Rewiring & $0.0282 \pm 0.0054$ & $0.0267$ & $0.0111$ & $0.0455$ & $0.0345$ & $\mathbf{0.3696}$ & $0.0376$ \\
    & T-ID swap & $\mathbf{0.9933} \pm 0.0149$ & $0.0171$ & $0.0111$ & $0.0147$ & $0.0133$ & $0.5317$ & $0.0340$ \\
    & Feature shuffling & $\mathbf{1.0000} \pm 0.0000$ & \textemdash & \textemdash & \textemdash & \textemdash & \textemdash & \textemdash \\
    & Feature shift & $\mathbf{0.9867} \pm 0.0182$ & \textemdash & \textemdash & \textemdash & \textemdash & \textemdash & \textemdash \\
    & F-ID swap & $\mathbf{1.0000} \pm 0.0000$ & \textemdash & \textemdash & \textemdash & \textemdash & \textemdash & \textemdash \\
\midrule
\multirow{11}{*}{\textbf{Wikipedia}} 
    & Time shuffling & $\mathbf{0.0584} \pm 0.0383$ & $0.0042$ & $0.0212$ & $0.0061$ & $0.0074$ & $0.0051$ & $0.0037$ \\
    & Time shift & $\mathbf{0.2188} \pm 0.1269$ & $0.0042$ & $0.0037$ & $0.0067$ & $0.0074$ & $0.0048$ & $0.0037$ \\
    & Link persistence & $\mathbf{0.4419} \pm 0.1606$ & $0.0042$ & $0.0037$ & $0.0070$ & $0.0057$ & $0.0051$ & $0.0067$ \\
    & Uniform densification & $0.0572 \pm 0.0188$ & $0.0042$ & $0.0037$ & $0.2066$ & $0.0074$ & $0.2234$ & $\mathbf{0.3333}$ \\
    & Structured densification & $\mathbf{0.0302} \pm 0.0079$ & $0.0042$ & $0.0037$ & $0.0104$ & $0.0061$ & $0.0048$ & $0.0036$ \\
    & Sparsification & $\mathbf{0.0504} \pm 0.0870$ & $0.0042$ & $0.0212$ & $0.0135$ & $0.0061$ & $0.0025$ & $0.0025$ \\
    & Rewiring & $0.0073 \pm 0.0018$ & $0.0042$ & $0.0037$ & $0.1088$ & $0.0075$ & $\mathbf{0.2101}$ & $0.0045$ \\
    & T-ID swap & $ 0.0103 \pm 0.0040$ & $0.0035$ & $0.0037$ & $0.0101$ & $0.0074$ & $\mathbf{0.2040}$ & $0.0038$ \\
    & Feature shuffling & $\mathbf{0.0630} \pm 0.0316$ & \textemdash & \textemdash & \textemdash & \textemdash & \textemdash & \textemdash \\
    & Feature shift & $\mathbf{0.0273} \pm 0.0089$ & \textemdash & \textemdash & \textemdash & \textemdash & \textemdash & \textemdash \\
    & F-ID swap & $\mathbf{0.0090} \pm 0.0029$ & \textemdash & \textemdash & \textemdash & \textemdash & \textemdash & \textemdash \\
\midrule
    \multirow{11}{*}{\textbf{Reddit}} 
    & Time shuffling & $\mathbf{0.9753} \pm 0.0246$ & $0.0041$ & $0.0037$ & $0.0073$ & $0.0051$ & $0.0084$ & $0.0088$ \\
    & Time shift & $\mathbf{0.9819} \pm 0.0261$ & $0.0036$ & $0.0037$ & $0.0064$ & $0.0051$ & $0.0085$ & $0.0088$ \\
    & Link persistence & $\mathbf{1.0000 }\pm 0.0000$ & $0.0038$ & $0.0037$ & $0.0040$ & $0.0545$ & $0.0085$ & $0.0273$ \\
    & Uniform densification & $\mathbf{1.0000} \pm 0.0000$ & $0.0037$ & $0.0037$ & $0.0115$ & $0.0056$ & $0.6099$ & $0.5025$ \\
    & Structured densification & $\mathbf{1.0000} \pm 0.0000$ & $0.0035$ & $0.0037$ & $0.0111$ & $0.0080$ & $0.0085$ & $0.0189$ \\
    & Sparsification & $\mathbf{0.0096} \pm 0.0007$ & $0.0053$ & $0.0037$ & $0.0075$ & $0.0047$ & $0.0030$ & $0.0034$ \\
    & Rewiring & $\mathbf{0.6564} \pm 0.0491$ & $0.0080$ & $0.0037$ & $0.0059$ & $0.0155$ & $0.3512$ & $0.0089$ \\
    & T-ID swap & $\mathbf{0.9654} \pm 0.0260$ & $0.0036$ & $0.0037$ & $0.0086$ & $0.0051$ & $0.4398$ & $0.6047$ \\
    & Feature shuffling & $\mathbf{0.9301} \pm 0.0278$ & \textemdash & \textemdash & \textemdash & \textemdash & \textemdash & \textemdash \\
    & Feature shift & $\mathbf{0.9782} \pm 0.0488$ & \textemdash & \textemdash & \textemdash & \textemdash & \textemdash & \textemdash \\
    & F-ID swap & $\mathbf{0.9800} \pm 0.0447$ & \textemdash & \textemdash & \textemdash & \textemdash & \textemdash & \textemdash \\
\midrule
\multirow{11}{*}{\textbf{MOOC}} 
    & Time shuffling & $\mathbf{0.0275} \pm 0.0066$ & $0.0038$ & $0.0081$ & $0.0045$ & $0.0121$ & $0.0034$ & $0.0040$ \\
    & Time shift & $\mathbf{0.0215} \pm 0.0050$ & $0.0038$ & $0.0081$ & $0.0043$ & $0.0121$ & $0.0034$ & $0.0040$ \\
    & Link persistence & $\mathbf{0.0891} \pm 0.0125$ & $0.0038$ & $0.0058$ & $0.0032$ & $0.0123$ & $0.0034$ & $0.0159$ \\
    & Uniform densification & $\mathbf{0.0477} \pm 0.0083$ & $0.0038$ & $0.0052$ & $0.0048$ & $0.0074$ & $0.0077$ & $0.0361$ \\
    & Structured densification & $\mathbf{0.0495} \pm 0.0079$ & $0.0038$ & $0.0058$ & $0.0063$ & $0.0147$ & $0.0034$ & $0.0058$ \\
    & Sparsification & $\mathbf{0.0495} \pm 0.0124$ & $0.0033$ & $0.0039$ & $0.0046$ & $0.0036$ & $0.0027$ & $0.0044$ \\
    & Rewiring & $\mathbf{0.0703} \pm 0.0619$ & $0.0038$ & $0.0081$ & $0.0041$ & $0.0119$ & $0.0041$ & $0.0039$ \\
    & T-ID swap & $0.0067 \pm 0.0022$ & $0.0033$ & $0.0081$ & $0.0042$ & $0.0121$ & $0.0036$ & $\mathbf{0.4021}$ \\
    & Feature shuffling & $\mathbf{0.0507} \pm 0.0349$ & \textemdash & \textemdash & \textemdash & \textemdash & \textemdash & \textemdash \\
    & Feature shift & $\mathbf{0.1476} \pm 0.1250$ & \textemdash & \textemdash & \textemdash & \textemdash & \textemdash & \textemdash \\
    & F-ID swap & $\mathbf{0.1041} \pm 0.0281$ & \textemdash & \textemdash & \textemdash & \textemdash & \textemdash & \textemdash \\
\midrule
\multirow{11}{*}{\textbf{LastFM}} 
    & Time shuffling & $\mathbf{0.1275} \pm 0.0164$ & $0.0106$ & $0.0105$ & $0.0188$ & $0.0164$ & $0.0137$ & $0.0120$ \\
    & Time shift & $\mathbf{0.1087} \pm 0.0096$ & $0.0106$ & $0.0105$ & $0.0187$ & $0.0164$ & $0.0135$ & $0.0120$ \\
    & Link persistence & $\mathbf{0.5331} \pm 0.1128$ & $0.0106$ & $0.0105$ & $0.0112$ & $0.0157$ & $0.0136$ & $0.2932$ \\
    & Uniform densification & $0.0720 \pm 0.0063$ & $0.0106$ & $0.0105$ & $0.0258$ & $0.0163$ & $0.6404$ & $\mathbf{0.7168}$ \\
    & Structured densification & $\mathbf{0.0710} \pm 0.0038$ & $0.0106$ & $0.0105$ & $0.0228$ & $0.0157$ & $0.0137$ & $0.0313$ \\
    & Sparsification & $0.0150 \pm 0.0008$ & $0.0106$ & $0.0105$ & $\mathbf{0.0186}$ & $0.0138$ & $0.0074$ & $0.0080$ \\
    & Rewiring & $0.0152 \pm 0.0008$ & $0.0106$ & $0.0105$ & $0.0217$ & $0.0187$ & $\mathbf{0.4456}$ & $0.0120$ \\
    & T-ID swap & $0.0165 \pm 0.0009$ & $0.0106$ & $0.0105$ & $0.0202$ & $0.0164$ & $0.4366$ & $\mathbf{1.0000}$ \\
    & Feature shuffling & $\mathbf{0.0445} \pm 0.0025$ & \textemdash & \textemdash & \textemdash & \textemdash & \textemdash & \textemdash \\
    & Feature shift & $\mathbf{0.0535} \pm 0.0058$ & \textemdash & \textemdash & \textemdash & \textemdash & \textemdash & \textemdash \\
    & F-ID swap & $\mathbf{0.0217} \pm 0.0009$ & \textemdash & \textemdash & \textemdash & \textemdash & \textemdash & \textemdash \\
\midrule
\multirow{11}{*}{\textbf{Enron}} 
    & Time shuffling & $0.0265 \pm 0.0017$ & $0.0129$ & $0.0323$ & $0.0286$ & $\mathbf{0.0434}$ & $0.0119$ & $0.0115$ \\
    & Time shift & $\mathbf{0.2316} \pm 0.0430$ & $0.0129$ & $0.0127$ & $0.0244$ & $0.0421$ & $0.0119$ & $0.0115$ \\
    & Link persistence & $\mathbf{0.3797} \pm 0.0939$ & $0.0129$ & $0.0127$ & $0.0239$ & $0.0459$ & $0.0119$ & $0.0143$ \\
    & Uniform densification & $\mathbf{0.1319} \pm 0.0633$  & $0.0129$ & $0.0323$ & $0.0343$ & $0.0387$ & $0.0182$ & $0.0147$ \\
    & Structured densification & $\mathbf{0.1601} \pm 0.0482$ & $0.0129$ & $0.0168$ & $0.0382$ & $0.0433$ & $0.0117$ & $0.0123$ \\
    & Sparsification & $0.0126 \pm 0.0008$ & $0.0109$ & $0.0127$ & $0.0171$ & $\mathbf{0.0321}$ & $0.0110$ & $0.0105$ \\
    & Rewiring & $0.0119 \pm 0.0006$ & $0.0129$ & $0.0127$ & $0.0267$ & $\mathbf{0.0422}$ & $0.0157$ & $0.0117$ \\
    & T-ID swap & $0.0134 \pm 0.0015$ & $0.0129$ & $0.0127$ & $0.0231$ & $\mathbf{0.0429}$ & $0.0132$ & $0.0119$ \\
    & Feature shuffling & $\mathbf{0.0123} \pm 0.0011$ & \textemdash & \textemdash & \textemdash & \textemdash & \textemdash & \textemdash \\
    & Feature shift & $\mathbf{0.1049} \pm 0.0157$ & \textemdash & \textemdash & \textemdash & \textemdash & \textemdash & \textemdash \\
    & F-ID swap & $\mathbf{0.0216} \pm 0.0113$ & \textemdash & \textemdash & \textemdash & \textemdash & \textemdash & \textemdash \\
\bottomrule
\end{tabular*}
\end{table*}

%% file: Sources/Appendix_ablation_tables.tex
\section{Detailed Ablation Tables}\label{sec:appendix-ablation_tables}

Tables \ref{appendix:ablation_table_roc} and \ref{appendix:ablation_table_pr} report the performance of MuTGN and the variants used in the ablation study across datasets and anomaly types, measured by ROC AUC and PR AUC, respectively.

\begin{table*}[!t]
\centering
\small
\setlength{\tabcolsep}{2.6pt}
\renewcommand{\arraystretch}{0.90}
\captionsetup{font=small,skip=3pt}
\caption{ROC AUC by dataset and anomaly type over five training seeds.}
\label{appendix:ablation_table_roc}
\begin{tabular*}{\textwidth}{@{\extracolsep{\fill}}llrrrr@{}}
\toprule
    \textbf{Dataset} & \textbf{Anomaly} & \textbf{MuTGN} & \textbf{No LayerNorm} & \textbf{No $\mathcal{E}$ norm.} & \textbf{No $\mathcal{E}$ sharing}   \\
\midrule
\multirow{11}{*}{\textbf{Synthetic}} 
    & Time shuffling & $\mathbf{1.0000} \pm 0.0000$ & $0.9851 \pm 0.0027$ & $0.9878 \pm 0.0095$ & $0.7455 \pm 0.0318$ \\
    & Time shift & $\mathbf{1.0000} \pm 0.0000$ & $0.9815 \pm 0.0049$ & $0.9891 \pm 0.0092$ & $0.6592 \pm 0.0979$ \\
    & Link persistence & $\mathbf{1.0000} \pm 0.0000$ & $\mathbf{1.0000} \pm 0.0000$ & $\mathbf{1.0000} \pm 0.0000$ & $0.6545 \pm 0.1362$ \\
    & Uniform densification & $\mathbf{0.9993} \pm 0.0005$ & $0.9769 \pm 0.0067$ & $0.9837 \pm 0.0098$ & $0.7764 \pm 0.0292$ \\
    & Structured densification & $\mathbf{0.9999} \pm 0.0002$ & $0.9805 \pm 0.0040$ & $0.9855 \pm 0.0087$ & $0.6399 \pm 0.1113$ \\
    & Sparsification & $\mathbf{0.9012} \pm 0.0214$ & $0.8703 \pm 0.0300$ & $0.7738 \pm 0.0186$ & $0.8394 \pm 0.0259$ \\
    & Rewiring & $0.7476 \pm 0.0459$ & $0.7147 \pm 0.0592$ & $0.7100 \pm 0.0715$ & $\mathbf{0.7961} \pm 0.0262$ \\
    & T-ID swap & $\mathbf{0.9999} \pm 0.0002$ & $0.9827 \pm 0.0073$ & $0.9745 \pm 0.0184$ & $0.9077 \pm 0.0355$ \\
    & Feature shuffling & $\mathbf{1.0000} \pm 0.0000$ & $0.9814 \pm 0.0029$ & $0.9753 \pm 0.0187$ & $0.7270 \pm 0.0711$ \\
    & Feature shift & $\mathbf{0.9998} \pm 0.0002$ & $0.9727 \pm 0.0047$ & $0.9853 \pm 0.0132$ & $0.6425 \pm 0.0706$ \\
    & F-ID swap & $\mathbf{1.0000} \pm 0.0000 $ & $\mathbf{1.0000} \pm 0.0000$ & $\mathbf{1.0000} \pm 0.0000$ & $\mathbf{1.0000} \pm 0.0000$ \\
\midrule
\multirow{11}{*}{\textbf{Wikipedia}} 
    & Time shuffling & $0.7324 \pm 0.0563$ & $\mathbf{0.7682} \pm 0.0440$ & $0.6252 \pm 0.0476$ & $0.4221 \pm 0.1615$ \\
    & Time shift & $0.9023 \pm 0.0226$ & $\mathbf{0.9358} \pm 0.0135$ & $0.6991 \pm 0.0247$ & $0.4196 \pm 0.1571$ \\
    & Link persistence & $0.9944 \pm 0.0047$ & $\mathbf{0.9976} \pm 0.0007$ & $0.9140 \pm 0.0198$ & $0.3375 \pm 0.0797$ \\
    & Uniform densification & $0.8979 \pm 0.0225$ & $\mathbf{0.9436} \pm 0.0191$ & $0.7607 \pm 0.0532$ & $0.5036 \pm 0.0758$ \\
    & Structured densification & $0.8641 \pm 0.0296$ & $\mathbf{0.9270} \pm 0.0191$ & $0.6806 \pm 0.0306$ & $0.2886 \pm 0.0966$ \\
    & Sparsification & $\mathbf{0.6226} \pm 0.0692$ & $0.5207 \pm 0.1099$ & $0.4619 \pm 0.1213$ & $0.5054 \pm 0.0387$ \\
    & Rewiring & $0.5892 \pm 0.0480$ & $0.4783 \pm 0.0395$ & $\mathbf{0.7451} \pm 0.0628$ & $0.5974 \pm 0.0488$ \\
    & T-ID swap & $0.6772 \pm 0.0522$ & $0.6197 \pm 0.0737$ & $\mathbf{0.8140} \pm 0.0393$ & $0.4340 \pm 0.0597$ \\
    & Feature shuffling & $0.7209 \pm 0.0414$ & $0.7899 \pm 0.0745$ & $\mathbf{0.8496} \pm 0.0596$ & $0.3978 \pm 0.1663$ \\
    & Feature shift & $0.7146 \pm 0.0547$ & $0.7869 \pm 0.0744$ & $\mathbf{0.8605} \pm 0.0629$ & $0.3970 \pm 0.1560$ \\
    & F-ID swap & $0.6000 \pm 0.0597$ & $0.6130 \pm 0.0532$ & $\mathbf{0.7678} \pm 0.0428$ & $0.3977 \pm 0.1502$ \\
\midrule
    \multirow{11}{*}{\textbf{Reddit}} 
    & Time shuffling & $\mathbf{0.9999} \pm 0.0001$ & $\mathbf{0.9999} \pm 0.0002$ & $0.9529 \pm 0.0204$ & $0.6551 \pm 0.0205$ \\
    & Time shift & $\mathbf{0.9999} \pm 0.0001$ & $\mathbf{0.9999} \pm 0.0002$ & $0.9541 \pm 0.0192$ & $0.6565 \pm 0.0298$ \\
    & Link persistence & $\mathbf{1.0000} \pm 0.0000$ & $\mathbf{1.0000} \pm 0.0000$ & $0.9838 \pm 0.0050$ & $0.7206 \pm 0.0265$ \\
    & Uniform densification & $\mathbf{1.0000} \pm 0.0000$ & $\mathbf{1.0000} \pm 0.0000$ & $0.9478 \pm 0.0234$ & $0.7116 \pm 0.0215$ \\
    & Structured densification & $\mathbf{1.0000} \pm 0.0000$ & $\mathbf{1.0000} \pm 0.0000$ & $0.9428 \pm 0.0237$ & $0.6489 \pm 0.0160$ \\
    & Sparsification & $0.7035 \pm 0.0241$ & $\mathbf{0.7318} \pm 0.0590$ & $0.6689 \pm 0.0564$ & $0.7195 \pm 0.0257$ \\
    & Rewiring & $0.9699 \pm 0.0083$ & $\mathbf{0.9749} \pm 0.0234$ & $0.9410 \pm 0.0282$ & $0.8123 \pm 0.0510$ \\
    & T-ID swap & $\mathbf{0.9999} \pm 0.0001$ & $\mathbf{0.9999} \pm 0.0000$ & $0.9696 \pm 0.0113$ & $0.8158 \pm 0.0271$ \\
    & Feature shuffling & $0.9996 \pm 0.0003$ & $\mathbf{0.9998} \pm 0.0002$ & $0.9731 \pm 0.0108$ & $0.6490 \pm 0.0150$ \\
    & Feature shift & $0.9998 \pm 0.0004$ & $\mathbf{0.9999} \pm 0.0001$ & $0.9727 \pm 0.0098$ & $0.6500 \pm 0.0190$ \\
    & F-ID swap & $0.9999 \pm 0.0003$ & $\mathbf{1.0000} \pm 0.0000$ & $0.9679 \pm 0.0116$ & $0.9692 \pm 0.0072$ \\
\midrule
\multirow{11}{*}{\textbf{MOOC}} 
    & Time shuffling & $\mathbf{0.7916} \pm 0.0122$ & $0.7755 \pm 0.0049$ & $0.7468 \pm 0.0318$ & $0.6165 \pm 0.0160$ \\
    & Time shift & $\mathbf{0.7830} \pm 0.0124$ & $0.7687 \pm 0.0082$ & $0.7569 \pm 0.0301$ & $0.6225 \pm 0.0172$ \\
    & Link persistence & $\mathbf{0.8416} \pm 0.0027$ & $0.8386 \pm 0.0021$ & $0.8223 \pm 0.0077$ & $0.5571 \pm 0.0399$ \\
    & Uniform densification & $\mathbf{0.7996} \pm 0.0055$ & $0.7921 \pm 0.0117$ & $0.7630 \pm 0.0087$ & $0.5901 \pm 0.0258$ \\
    & Structured densification & $\mathbf{0.8048} \pm 0.0061$ & $0.7931 \pm 0.0079$ & $0.7593 \pm 0.0313$ & $0.5807 \pm 0.0126$ \\
    & Sparsification & $0.6855 \pm 0.0215$ & $0.1812 \pm 0.0618$ & $\mathbf{0.7122} \pm 0.0284$ & $0.5741 \pm 0.1020$ \\
    & Rewiring & $0.7552 \pm 0.0742$ & $\mathbf{0.8032} \pm 0.0116$ & $0.5534 \pm 0.0932$ & $0.6092 \pm 0.0148$ \\
    & T-ID swap & $0.4540 \pm 0.0778$ & $0.4787 \pm 0.0836$ & $0.4898 \pm 0.0903$ & $\mathbf{0.7311} \pm 0.0187$ \\
    & Feature shuffling & $0.7259 \pm 0.0776$ & $\mathbf{0.8347} \pm 0.0046$ & $0.5992 \pm 0.0402$ & $0.6238 \pm 0.0217$ \\
    & Feature shift & $0.7211 \pm 0.0480$ & $\mathbf{0.8259} \pm 0.0130$ & $0.6357 \pm 0.0607$ & $0.6196 \pm 0.0133$ \\
    & F-ID swap & $0.6585 \pm 0.0194$ & $0.7763 \pm 0.0714$ & $0.7363 \pm 0.0688$ & $\mathbf{0.7989} \pm 0.0238$ \\
\midrule
\multirow{11}{*}{\textbf{LastFM}} 
    & Time shuffling & $\mathbf{0.9581} \pm 0.0069$ & $0.6640 \pm 0.1741$ & $0.6385 \pm 0.1531$ & $0.6067 \pm 0.0599$ \\
    & Time shift & $\mathbf{0.9505} \pm 0.0059$ & $0.6460 \pm 0.1767$ & $0.6261 \pm 0.1671$ & $0.5794 \pm 0.0546$ \\
    & Link persistence & $\mathbf{0.9930} \pm 0.0038$ & $0.6601 \pm 0.1886$ & $0.6967 \pm 0.1604$ & $0.5614 \pm 0.0296$ \\
    & Uniform densification & $\mathbf{0.9165} \pm 0.0120$ & $0.7128 \pm 0.1137$ & $0.6619 \pm 0.0820$ & $0.6355 \pm 0.0137$ \\
    & Structured densification & $\mathbf{0.9166} \pm 0.0091$ & $0.6225 \pm 0.1665$ & $0.5967 \pm 0.1443$ & $0.5641 \pm 0.0458$ \\
    & Sparsification & $0.5923 \pm 0.0258$ & $0.6126 \pm 0.0557$ & $0.5176 \pm 0.0631$ & $\mathbf{0.6286} \pm 0.0667$ \\
    & Rewiring & $\mathbf{0.5758} \pm 0.0250$ & $0.5559 \pm 0.0399$ & $0.4769 \pm 0.0436$ & $0.5690 \pm 0.0466$ \\
    & T-ID swap & $\mathbf{0.6111} \pm 0.0285$ & $0.5574 \pm 0.0822$ & $0.4605 \pm 0.0260$ & $0.5408 \pm 0.0533$ \\
    & Feature shuffling & $\mathbf{0.8545} \pm 0.0058$ & $0.6306 \pm 0.1384$ & $0.6532 \pm 0.0856$ & $0.5801 \pm 0.0460$ \\
    & Feature shift & $\mathbf{0.8910} \pm 0.0138$ & $0.6320 \pm 0.1512$ & $0.6594 \pm 0.0854$ & $0.5760 \pm 0.0431$ \\
    & F-ID swap & $\mathbf{0.6996} \pm 0.0180$ & $0.5979 \pm 0.0812$ & $0.5054 \pm 0.0831$ & $0.5827 \pm 0.0484$ \\
\midrule
\multirow{11}{*}{\textbf{Enron}} 
    & Time shuffling & $\mathbf{0.6285} \pm 0.0205$ & $0.3640 \pm 0.1180$ & $0.5336 \pm 0.0476$ & $0.3632 \pm 0.0207$ \\
    & Time shift & $\mathbf{0.6865} \pm 0.0138$ & $0.4955 \pm 0.1057$ & $0.6166 \pm 0.0248$ & $0.3875 \pm 0.0174$ \\
    & Link persistence & $\mathbf{0.7945} \pm 0.0167$ & $0.5512 \pm 0.1411$ & $0.6411 \pm 0.0397$ & $0.3169 \pm 0.0163$ \\
    & Uniform densification & $\mathbf{0.7233} \pm 0.0182$ & $0.4577 \pm 0.1343$ & $0.6053 \pm 0.0220$ & $0.3842 \pm 0.0108$ \\
    & Structured densification & $\mathbf{0.7390} \pm 0.0119$ & $0.4680 \pm 0.1391$ & $0.6030 \pm 0.0172$ & $0.3158 \pm 0.0140$ \\
    & Sparsification & $0.4203 \pm 0.0340$ & $0.3477 \pm 0.0510$ & $\mathbf{0.4587} \pm 0.0266$ & $0.3379 \pm 0.0094$ \\
    & Rewiring & $0.3932 \pm 0.0266$ & $0.3376 \pm 0.0512$ & $\mathbf{0.4785} \pm 0.0283$ & $0.4768 \pm 0.0138$ \\
    & T-ID swap & $0.4547 \pm 0.0568$ & $0.3423 \pm 0.0546$ & $\mathbf{0.5452} \pm 0.0380$ & $0.4749 \pm 0.0110$\\
    & Feature shuffling & $0.4059 \pm 0.0486$ & $0.3220 \pm 0.0672$ & $\mathbf{0.5302} \pm 0.0548$ & $0.3553 \pm 0.0126$ \\
    & Feature shift & $0.6184 \pm 0.0134$ & $0.4991 \pm 0.0834$ & $\mathbf{0.6193} \pm 0.0524$ & $0.3439 \pm 0.0069$ \\
    & F-ID swap & $0.5227 \pm 0.0261$ & $0.4187 \pm 0.0665$ & $\mathbf{0.6022} \pm 0.0239$ & $0.4344 \pm 0.0114$ \\
\bottomrule
\end{tabular*}
\end{table*}

\begin{table*}[!t]
\centering
\small
\setlength{\tabcolsep}{2.6pt}
\renewcommand{\arraystretch}{0.90}
\captionsetup{font=small,skip=3pt}
\caption{PR AUC by dataset and anomaly type over five training seeds.}
\label{appendix:ablation_table_pr}
\begin{tabular*}{\textwidth}{@{\extracolsep{\fill}}llrrrr@{}}
\toprule
    \textbf{Dataset} & \textbf{Anomaly} & \textbf{MuTGN} & \textbf{No LayerNorm} & \textbf{No $\mathcal{E}$ norm.} & \textbf{No $\mathcal{E}$ sharing}   \\
\midrule
\multirow{11}{*}{\textbf{Synthetic}} 
    & Time shuffling & $\mathbf{1.0000} \pm 0.0000$ & $0.3583 \pm 0.0633$ & $0.6168 \pm 0.1587$ & $0.0710 \pm 0.0418$ \\
    & Time shift & $\mathbf{1.0000} \pm 0.0000$ & $0.2863 \pm 0.0604$ & $0.6636 \pm 0.1452$ & $0.0371 \pm 0.0401$ \\
    & Link persistence & $\mathbf{1.0000} \pm 0.0000$ & $\mathbf{1.0000} \pm 0.0000$ & $\mathbf{1.0000} \pm 0.0000$ & $0.0694 \pm 0.1137$ \\
    & Uniform densification & $\mathbf{0.9555} \pm 0.0290$ & $0.3734 \pm 0.1154$ & $0.6497 \pm 0.1523$ & $0.0814 \pm 0.0853$ \\
    & Structured densification & $\mathbf{0.9933} \pm 0.0149$ & $0.3740 \pm 0.1335$ & $0.5486 \pm 0.0959$ & $0.0664 \pm 0.1108$ \\
    & Sparsification & $\mathbf{0.4694} \pm 0.1211$ & $0.3589 \pm 0.0577$ & $0.1045 \pm 0.0697$ & $0.4313 \pm 0.0198$ \\
    & Rewiring & $0.0282 \pm 0.0054$ & $0.0298 \pm 0.0063$ & $0.0382 \pm 0.0167$ & $\mathbf{0.1312} \pm 0.0337$ \\
    & T-ID swap & $\mathbf{0.9933} \pm 0.0149$ & $0.3411 \pm 0.0860$ & $0.5124 \pm 0.0990$ & $0.3623 \pm 0.1341$ \\
    & Feature shuffling & $\mathbf{1.0000} \pm 0.0000$ & $0.3171 \pm 0.0456$ & $0.6266 \pm 0.0933$ & $0.0875 \pm 0.0954$ \\
    & Feature shift & $\mathbf{0.9867} \pm 0.0182$ & $0.2458 \pm 0.0407$ & $0.6550 \pm 0.1632$ & $0.0324 \pm 0.0311$ \\
    & F-ID swap & $\mathbf{1.0000} \pm 0.0000$ & $\mathbf{1.0000} \pm 0.0000$ & $\mathbf{1.0000} \pm 0.0000$ & $\mathbf{1.0000} \pm 0.0000$ \\
\midrule
\multirow{11}{*}{\textbf{Wikipedia}} 
    & Time shuffling & $0.0584 \pm 0.0383$ & $\mathbf{0.0887} \pm 0.0489$ & $0.0084 \pm 0.0008$ & $0.0040 \pm 0.0013$ \\
    & Time shift & $0.2188 \pm 0.1269$ & $\mathbf{0.3046} \pm 0.1109$ & $0.0101 \pm 0.0016$ & $0.0040 \pm 0.0012$ \\
    & Link persistence & $0.4419 \pm 0.1606$ & $\mathbf{0.5709} \pm 0.1217$ & $0.0260 \pm 0.0060$ & $0.0034 \pm 0.0007$ \\
    & Uniform densification & $0.0572 \pm 0.0188$ & $\mathbf{0.0789} \pm 0.0121$ & $0.0182 \pm 0.0070$ & $0.0504 \pm 0.0853$ \\
    & Structured densification & $0.0302 \pm 0.0079$ & $\mathbf{0.0603} \pm 0.0135$ & $0.0083 \pm 0.0015$ & $0.0042 \pm 0.0030$ \\
    & Sparsification & $\mathbf{0.0504} \pm 0.0870$ & $0.0098 \pm 0.0053$ & $0.0049 \pm 0.0026$ & $0.0051 \pm 0.0009$ \\
    & Rewiring & $0.0073 \pm 0.0018$ & $0.0052 \pm 0.0004$ & $\mathbf{0.0105} \pm 0.0034$ & $0.0079 \pm 0.0027$ \\
    & T-ID swap & $0.0103 \pm 0.0040$ & $0.0102 \pm 0.0033$ & $\mathbf{0.0144} \pm 0.0047$ & $0.0038 \pm 0.0004$ \\
    & Feature shuffling & $0.0630 \pm 0.0316$ & $\mathbf{0.1046} \pm 0.0806$ & $0.0256 \pm 0.0128$ & $0.0039 \pm 0.0013$ \\
    & Feature shift & $0.0273 \pm 0.0089$ & $\mathbf{0.0509} \pm 0.0194$ & $0.0274 \pm 0.0119$ & $0.0038 \pm 0.0011$ \\
    & F-ID swap & $0.0090 \pm 0.0029$ & $\mathbf{0.0247} \pm 0.0123$ & $0.0160 \pm 0.0054$ & $0.0038 \pm 0.0010$ \\
\midrule
    \multirow{11}{*}{\textbf{Reddit}} 
    & Time shuffling & $0.9753 \pm 0.0246$ & $\mathbf{0.9772} \pm 0.0313$ & $0.0620 \pm 0.0164$ & $0.0072 \pm 0.0005$ \\
    & Time shift & $\mathbf{0.9819} \pm 0.0261$ & $0.9772 \pm 0.0313$ & $0.0645 \pm 0.0176$ & $0.0073 \pm 0.0007$ \\
    & Link persistence & $\mathbf{1.0000} \pm 0.0000$ & $\mathbf{1.0000} \pm 0.0000$ & $0.1214 \pm 0.0267$ & $0.0094 \pm 0.0023$ \\
    & Uniform densification & $\mathbf{1.0000} \pm 0.0000$ & $0.9933 \pm 0.0149$ & $0.0462 \pm 0.0129$ & $0.0088 \pm 0.0009$ \\
    & Structured densification & $\mathbf{1.0000} \pm 0.0000$ & $\mathbf{1.0000} \pm 0.0000$ & $0.0450 \pm 0.0118$ & $0.0072 \pm 0.0005$ \\
    & Sparsification & $0.0096 \pm 0.0007$ & $\mathbf{0.0552} \pm 0.0879$ & $0.0076 \pm 0.0015$ & $0.0093 \pm 0.0018$ \\
    & Rewiring & $0.6564 \pm 0.0491$ & $\mathbf{0.6579} \pm 0.0688$ & $0.0439 \pm 0.0126$ & $0.0137 \pm 0.0030$ \\
    & T-ID swap & $0.9654 \pm 0.0260$ & $\mathbf{0.9734} \pm 0.0149$ & $0.0777 \pm 0.0258$ & $0.0122 \pm 0.0015$ \\
    & Feature shuffling & $0.9301 \pm 0.0278$ & $\mathbf{0.9622} \pm 0.0353$ & $0.0859 \pm 0.0227$ & $0.0072 \pm 0.0006$ \\
    & Feature shift & $0.9782 \pm 0.0488$ & $\mathbf{0.9819} \pm 0.0261$ & $0.0812 \pm 0.0198$ & $0.0072 \pm 0.0005$ \\
    & F-ID swap & $0.9800 \pm 0.0447$ & $\mathbf{1.0000} \pm 0.0000$ & $0.0740 \pm 0.0211$ & $0.1797 \pm 0.0319$ \\
\midrule
\multirow{11}{*}{\textbf{MOOC}} 
    & Time shuffling & $0.0275 \pm 0.0066$ & $0.0177 \pm 0.0021$ & $0.0141 \pm 0.0035$ & $\mathbf{0.0633} \pm 0.0823$ \\
    & Time shift & $0.0215 \pm 0.0050$ & $0.0162 \pm 0.0042$ & $0.0162 \pm 0.0035$ & $\mathbf{0.0677} \pm 0.0821$ \\
    & Link persistence & $\mathbf{0.0891} \pm 0.0125$ & $0.0763 \pm 0.0073$ & $0.0406 \pm 0.0080$ & $0.0155 \pm 0.0178$ \\
    & Uniform densification & $\mathbf{0.0477} \pm 0.0083$ & $0.0350 \pm 0.0065$ & $0.0194 \pm 0.0048$ & $0.0119 \pm 0.0040$ \\
    & Structured densification & $\mathbf{0.0495} \pm 0.0079$ & $0.0392 \pm 0.0036$ & $0.0212 \pm 0.0066$ & $0.0142 \pm 0.0109$ \\
    & Sparsification & $\mathbf{0.0495} \pm 0.0124$ & $0.0027 \pm 0.0002$ & $0.0450 \pm 0.0116$ & $0.0185 \pm 0.0112$ \\
    & Rewiring & $0.0703 \pm 0.0619$ & $\mathbf{0.0790} \pm 0.0417$ & $0.0073 \pm 0.0037$ & $0.0171 \pm 0.0036$ \\
    & T-ID swap & $0.0067 \pm 0.0022$ & $0.0062 \pm 0.0015$ & $0.0065 \pm 0.0025$ & $\mathbf{0.0279} \pm 0.0112$ \\
    & Feature shuffling & $0.0507 \pm 0.0349$ & $\mathbf{0.0752} \pm 0.0209$ & $0.0102 \pm 0.0025$ & $0.0391 \pm 0.0212$ \\
    & Feature shift & $\mathbf{0.1476} \pm 0.1250$ & $0.0854 \pm 0.0277$ & $0.0103 \pm 0.0018$ & $0.0457 \pm 0.0158$ \\
    & F-ID swap & $\mathbf{0.1041} \pm 0.0281$ & $0.0497 \pm 0.0094$ & $0.0703 \pm 0.0267$ & $0.0921 \pm 0.0754$ \\
\midrule
\multirow{11}{*}{\textbf{LastFM}} 
    & Time shuffling & $\mathbf{0.1275} \pm 0.0164$ & $0.0587 \pm 0.0932$ & $0.0194 \pm 0.0083$ & $0.0167 \pm 0.0016$ \\
    & Time shift & $\mathbf{0.1087} \pm 0.0096$ & $0.0367 \pm 0.0457$ & $0.0191 \pm 0.0085$ & $0.0159 \pm 0.0013$ \\
    & Link persistence & $\mathbf{0.5331} \pm 0.1128$ & $0.1290 \pm 0.2517$ & $0.0257 \pm 0.0149$ & $0.0146 \pm 0.0008$ \\
    & Uniform densification & $\mathbf{0.0720} \pm 0.0063$ & $0.0288 \pm 0.0202$ & $0.0191 \pm 0.0046$ & $0.0182 \pm 0.0007$ \\
    & Structured densification & $\mathbf{0.0710} \pm 0.0038$ & $0.0248 \pm 0.0225$ & $0.0169 \pm 0.0059$ & $0.0146 \pm 0.0008$ \\
    & Sparsification & $0.0150 \pm 0.0008$ & $\mathbf{0.0242} \pm 0.0071$ & $0.0129 \pm 0.0013$ & $0.0223 \pm 0.0045$ \\
    & Rewiring & $\mathbf{0.0152} \pm 0.0008$ & $0.0146 \pm 0.0008$ & $0.0118 \pm 0.0007$ & $0.0151 \pm 0.0010$ \\
    & T-ID swap & $\mathbf{0.0165} \pm 0.0009$ & $0.0151 \pm 0.0023$ & $0.0114 \pm 0.0003$ & $0.0144 \pm 0.0012$ \\
    & Feature shuffling & $\mathbf{0.0445} \pm 0.0025$ & $0.0216 \pm 0.0127$ & $0.0187 \pm 0.0046$ & $0.0159 \pm 0.0009$ \\
    & Feature shift & $\mathbf{0.0535} \pm 0.0058$ & $0.0236 \pm 0.0178$ & $0.0191 \pm 0.0049$ & $0.0158 \pm 0.0010$ \\
    & F-ID swap & $\mathbf{0.0217} \pm 0.0009$ & $0.0172 \pm 0.0027$ & $0.0126 \pm 0.0022$ & $0.0161 \pm 0.0009$ \\
\midrule
\multirow{11}{*}{\textbf{Enron}} 
    & Time shuffling & $\mathbf{0.0265} \pm 0.0017$ & $0.0122 \pm 0.0036$ & $0.0244 \pm 0.0049$ & $0.0113 \pm 0.0004$ \\
    & Time shift & $\mathbf{0.2316} \pm 0.0430$ & $0.0285 \pm 0.0290$ & $0.0381 \pm 0.0089$ & $0.0119 \pm 0.0005$ \\
    & Link persistence & $\mathbf{0.3797} \pm 0.0939$ & $0.0412 \pm 0.0526$ & $0.0329 \pm 0.0044$ & $0.0106 \pm 0.0003$ \\
    & Uniform densification & $\mathbf{0.1319} \pm 0.0633$ & $0.0196 \pm 0.0165$ & $0.0284 \pm 0.0052$ & $0.0118 \pm 0.0003$ \\
    & Structured densification & $\mathbf{0.1601} \pm 0.0482$ & $0.0195 \pm 0.0131$ & $0.0283 \pm 0.0059$ & $0.0105 \pm 0.0003$ \\
    & Sparsification & $0.0126 \pm 0.0008$ & $0.0112 \pm 0.0008$ & $\mathbf{0.0182} \pm 0.0033$ & $0.0109 \pm 0.0002$ \\
    & Rewiring & $0.0119 \pm 0.0006$ & $0.0110 \pm 0.0008$ & $\mathbf{0.0204} \pm 0.0051$ & $0.0159 \pm 0.0013$ \\
    & T-ID swap & $0.0134 \pm 0.0015$ & $0.0110 \pm 0.0010$ & $\mathbf{0.0236} \pm 0.0031$ & $0.0162 \pm 0.0014$ \\
    & Feature shuffling & $0.0123 \pm 0.0011$ & $0.0107 \pm 0.0009$ & $\mathbf{0.0206} \pm 0.0016$ & $0.0112 \pm 0.0002$ \\
    & Feature shift & $\mathbf{0.1049} \pm 0.0157$ & $0.0240 \pm 0.0156$ & $0.0279 \pm 0.0076$ & $0.0110 \pm 0.0001$ \\
    & F-ID swap & $0.0216 \pm 0.0113$ & $0.0136 \pm 0.0020$ & $\mathbf{0.0307} \pm 0.0047$ & $0.0136 \pm 0.0007$ \\
\bottomrule

\end{tabular*}
\end{table*}

%% file: biblio.bib
@article{bhatia.2022.midas,
    author = {Bhatia, Siddharth and Liu, Rui and Hooi, Bryan and Yoon, Minji and Shin, Kijung and Faloutsos, Christos},
    title = {Real-Time Anomaly Detection in Edge Streams},
    year = {2022},
    issue_date = {August 2022},
    publisher = {Association for Computing Machinery},
    address = {New York, NY, USA},
    volume = {16},
    number = {4},
    issn = {1556-4681},
    url = {https://doi.org/10.1145/3494564},
    doi = {10.1145/3494564},
    journal = {ACM Trans. Knowl. Discov. Data},
    month = jan,
    articleno = {75},
    numpages = {22}
}

@inproceedings{chang.2021.ffade,
  title={F-fade: Frequency factorization for anomaly detection in edge streams},
  author={Chang, Yen-Yu and Li, Pan and Sosic, Rok and Afifi, MH and Schweighauser, Marco and Leskovec, Jure},
  booktitle={Proceedings of the 14th ACM international conference on web search and data mining},
  pages={589--597},
  year={2021}
}

@article{liu.2023.taddy,
   title={Anomaly Detection in Dynamic Graphs via Transformer},
   volume={35},
   ISSN={2326-3865},
   url={http://dx.doi.org/10.1109/TKDE.2021.3124061},
   DOI={10.1109/tkde.2021.3124061},
   number={12},
   journal={IEEE Transactions on Knowledge and Data Engineering},
   publisher={Institute of Electrical and Electronics Engineers (IEEE)},
   author={Liu, Yixin and Pan, Shirui and Wang, Yu Guang and Xiong, Fei and Wang, Liang and Chen, Qingfeng and Lee, Vincent CS},
   year={2023},
   month=dec, pages={12081–12094} }

@inproceedings{eswaran.2018.sedanspot,
  title={Sedanspot: Detecting anomalies in edge streams},
  author={Eswaran, Dhivya and Faloutsos, Christos},
  booktitle={2018 IEEE International conference on data mining (ICDM)},
  pages={953--958},
  year={2018},
  organization={IEEE}
}

@inproceedings{liu.2024.anomalyllm,
  title={AnomalyLLM: Few-shot anomaly edge detection for dynamic graphs using large language models},
  author={Liu, Shuo and Yao, Di and Fang, Lanting and Li, Zhetao and Li, Wenbin and Feng, Kaiyu and Ji, Xiaowen and Bi, Jingping},
  booktitle={2024 IEEE International Conference on Data Mining (ICDM)},
  pages={785--790},
  year={2024},
  organization={IEEE}
}

@article{liu.2025.comedy,
  title={COMEDY: Continuous-Time Anomalous Edge Detection in Dynamic Networks},
  author={Liu, Jie and Liu, Jiamou and Zhao, Kaiqi and Chen, Wu},
  journal={IEEE Transactions on Network Science and Engineering},
  year={2025},
  publisher={IEEE}
}

@article{bansal.2022.dsedn,
  title={Density-based structural embedding for anomaly detection in dynamic networks},
  author={Bansal, Monika and Sharma, Dolly},
  journal={Neurocomputing},
  volume={500},
  pages={724--740},
  year={2022},
  publisher={Elsevier}
}

@article{jiao.2021.geabs,
  title={Generative evolutionary anomaly detection in dynamic networks},
  author={Jiao, Pengfei and Li, Tianpeng and Xie, Yingjie and Wang, Yinghui and Wang, Wenjun and He, Dongxiao and Wu, Huaming},
  journal={IEEE Transactions on Knowledge and Data Engineering},
  volume={35},
  number={12},
  pages={12234--12248},
  year={2021},
  publisher={IEEE}
}

@inproceedings{lee.2024.slade,
  title={Slade: Detecting dynamic anomalies in edge streams without labels via self-supervised learning},
  author={Lee, Jongha and Kim, Sunwoo and Shin, Kijung},
  booktitle={Proceedings of the 30th ACM SIGKDD Conference on Knowledge Discovery and Data Mining},
  pages={1506--1517},
  year={2024}
}

@inproceedings{chakraborty.2025.flex,
author = {Chakraborty, Apu and Kumar, Anshul and Gupta, Gagan Raj},
title = {Flex-GAD: Flexible Graph Anomaly Detection},
year = {2026},
isbn = {9798400723551},
publisher = {Association for Computing Machinery},
address = {New York, NY, USA},
url = {https://doi.org/10.1145/3799830.3799843},
doi = {10.1145/3799830.3799843},
booktitle = {Proceedings of the 13th ACM IKDD International Conference on Data Science},
pages = {115–123},
numpages = {9},
location = {Pune, India},
series = {CODS '25}
}

@article{liu.2024.stripe,
  title={Detecting anomalies in dynamic graphs via memory enhanced normality},
  author={Liu, Jie and Shang, Xuequn and Han, Xiaolin and Zheng, Kai and Yin, Hongzhi},
  journal={arXiv preprint arXiv:2403.09039},
  year={2024}
}

@inproceedings{yoon.2019.anomrank,
  title={Fast and accurate anomaly detection in dynamic graphs with a two-pronged approach},
  author={Yoon, Minji and Hooi, Bryan and Shin, Kijung and Faloutsos, Christos},
  booktitle={Proceedings of the 25th ACM SIGKDD international conference on knowledge discovery \& data mining},
  pages={647--657},
  year={2019}
}

@inproceedings{guo.2022.dynanom,
   series={KDD ’22},
   title={Subset Node Anomaly Tracking over Large Dynamic Graphs},
   url={http://dx.doi.org/10.1145/3534678.3539389},
   DOI={10.1145/3534678.3539389},
   booktitle={Proceedings of the 28th ACM SIGKDD Conference on Knowledge Discovery and Data Mining},
   publisher={ACM},
   author={Guo, Xingzhi and Zhou, Baojian and Skiena, Steven},
   year={2022},
   month=aug, pages={475–485},
   collection={KDD ’22} 
}

@inproceedings{eswaran.2018.spotlight,
  title={Spotlight: Detecting anomalies in streaming graphs},
  author={Eswaran, Dhivya and Faloutsos, Christos and Guha, Sudipto and Mishra, Nina},
  booktitle={Proceedings of the 24th ACM SIGKDD international conference on knowledge discovery \& data mining},
  pages={1378--1386},
  year={2018}
}

@inproceedings{huang.2020.lad,
   series={KDD ’20},
   title={Laplacian Change Point Detection for Dynamic Graphs},
   url={http://dx.doi.org/10.1145/3394486.3403077},
   DOI={10.1145/3394486.3403077},
   booktitle={Proceedings of the 26th ACM SIGKDD International Conference on Knowledge Discovery \&amp; Data Mining},
   publisher={ACM},
   author={Huang, Shenyang and Hitti, Yasmeen and Rabusseau, Guillaume and Rabbany, Reihaneh},
   year={2020},
   month=aug, pages={349–358},
   collection={KDD ’20}
}

@inproceedings{mcneil.2023.cadence,
  title={Cadence: Community-aware detection of dynamic network states},
  author={McNeil, Maxwell and Mattsson, Carolina and Takes, Frank W and Bogdanov, Petko},
  booktitle={Proceedings of the 2023 SIAM International Conference on Data Mining (SDM)},
  pages={1--9},
  year={2023},
  organization={SIAM}
}

@article{luo.2025.graph,
  title={Graph online change point detection based on fr{\'e}chet statistics},
  author={Luo, Rui and Bao, Jie and So, Hing Cheung and Cao, Suqun and Xu, Mengqiao},
  journal={Neurocomputing},
  pages={131874},
  year={2025},
  publisher={Elsevier}
}

@article{jiang.2022.tsad,
  title={Two-stage anomaly detection algorithm via dynamic community evolution in temporal graph},
  author={Jiang, Yan and Liu, Guannan},
  journal={Applied Intelligence},
  volume={52},
  number={11},
  pages={12222--12240},
  year={2022},
  publisher={Springer}
}

@inproceedings{tasnadi.2022.new,
  title={A new type of anomaly detection problem in dynamic graphs: An ant colony optimization approach},
  author={Tasn{\'a}di, Zolt{\'a}n and Gask{\'o}, No{\'e}mi},
  booktitle={International Conference on Bioinspired Optimization Methods and Their Applications},
  pages={46--53},
  year={2022},
  organization={Springer}
}

@inproceedings{bautista.2024.mad,
author = {Bautista, Esteban and Brisson, Laurent and Bothorel, C{\'e}cile and Smits, Gr{\'e}gory},
title = {MAD: Multi-Scale Anomaly Detection in Link Streams},
year = {2024},
isbn = {9798400703713},
publisher = {Association for Computing Machinery},
address = {New York, NY, USA},
url = {https://doi.org/10.1145/3616855.3635834},
doi = {10.1145/3616855.3635834},
booktitle = {Proceedings of the 17th ACM International Conference on Web Search and Data Mining},
pages = {38–46},
numpages = {9},
location = {Merida, Mexico},
series = {WSDM '24}
}

@inproceedings{sun.2006.dta,
    title = "Beyond streams and graphs: Dynamic tensor analysis",
    author = "Jimeng Sun and Dacheng Tao and Christos Faloutsos",
    year = "2006",
    month = oct,
    day = "16",
    language = "English (US)",
    isbn = "1595933395",
    series = "Proceedings of the ACM SIGKDD International Conference on Knowledge Discovery and Data Mining",
    pages = "374--383",
    booktitle = "KDD 2006",
    note = "KDD 2006: 12th ACM SIGKDD International Conference on Knowledge Discovery and Data Mining ; Conference date: 20-08-2006 Through 23-08-2006",
}

@inproceedings{koutra.2012.tensorsplat,
  author    = {Danai Koutra and Evangelos E. Papalexakis and Christos Faloutsos},
  title     = {TENSORSPLAT: Spotting Latent Anomalies in Time},
  booktitle = {2012 IEEE 12th International Conference on Data Mining Workshops (ICDMW)},
  year      = {2012},
  pages     = {559--568},
  doi       = {10.1109/ICDMW.2012.43}
}

@inproceedings{leggas.2021.knowledge,
  title={Knowledge-guided Tensor Decomposition for Baselining and Anomaly Detection},
  author={Leggas, Dimitri and Coley, Christopher and Ranadive, Teresa},
  booktitle={2021 IEEE High Performance Extreme Computing Conference (HPEC)},
  pages={1--7},
  year={2021},
  organization={IEEE}
}

@article{fernandes.2021.wintended,
    author = {Fernandes, Sofia and Fanaee-T, Hadi and Gama, Jo\~{a}o and Ti\v{s}ljari\'{c}, Leo and \v{S}muc, Tomislav},
    title = {WINTENDED: WINdowed TENsor decomposition for Densification Event Detection in time-evolving networks},
    year = {2021},
    issue_date = {Feb 2023},
    publisher = {Kluwer Academic Publishers},
    address = {USA},
    volume = {112},
    number = {2},
    issn = {0885-6125},
    url = {https://doi.org/10.1007/s10994-021-05979-8},
    doi = {10.1007/s10994-021-05979-8},
    journal = {Mach. Learn.},
    month = may,
    pages = {459–481},
    numpages = {23}
}

@ARTICLE{xu.2019.swtf,
  author={Xu, Ming and Wu, Jianping and Wang, Haohan and Cao, Mengxin},
  journal={IEEE Transactions on Intelligent Transportation Systems}, 
  title={Anomaly Detection in Road Networks Using Sliding-Window Tensor Factorization}, 
  year={2019},
  volume={20},
  number={12},
  pages={4704-4713},
  doi={10.1109/TITS.2019.2941649}
}

@inproceedings{king.2023.edgetorrent,
  title={EdgeTorrent: Real-time temporal graph representations for intrusion detection},
  author={King, Isaiah J and Shu, Xiaokui and Jang, Jiyong and Eykholt, Kevin and Lee, Taesung and Huang, H Howie},
  booktitle={Proceedings of the 26th International Symposium on Research in Attacks, Intrusions and Defenses},
  pages={77--91},
  year={2023}
}

@inproceedings{justin.2026.madbgp,
author = {Loye, Justin and Bautista, Esteban and Fontugne, Romain},
title = {When BGP Goes MAD: Multi-Scale Anomaly Detection in Internet Routing},
year = {2026},
isbn = {9798400723278},
publisher = {Association for Computing Machinery},
address = {New York, NY, USA},
url = {https://doi.org/10.1145/3777912.3809143},
doi = {10.1145/3777912.3809143},
booktitle = {Proceedings of the 2026 ACM Internet Measurement Conference},
pages = {855–871},
numpages = {17},
location = {Karlsruhe, Germany},
series = {IMC '26}
}

@article{khodabandehlou.2024.fifraud,
  title={FiFrauD: unsupervised financial fraud detection in dynamic graph streams},
  author={Khodabandehlou, Samira and Golpayegani, Alireza Hashemi},
  journal={ACM Transactions on Knowledge Discovery from Data},
  volume={18},
  number={5},
  pages={1--29},
  year={2024},
  publisher={ACM New York, NY}
}

@article{zhao.2019.tgcn,
  title={T-GCN: A temporal graph convolutional network for traffic prediction},
  author={Zhao, Ling and Song, Yujiao and Zhang, Chao and Liu, Yu and Wang, Pu and Lin, Tao and Deng, Min and Li, Haifeng},
  journal={IEEE transactions on intelligent transportation systems},
  volume={21},
  number={9},
  pages={3848--3858},
  year={2019},
  publisher={IEEE}
}

@inproceedings{kumar.2019.jodie,
  title={Predicting dynamic embedding trajectory in temporal interaction networks},
  author={Kumar, Srijan and Zhang, Xikun and Leskovec, Jure},
  booktitle={Proceedings of the 25th ACM SIGKDD international conference on knowledge discovery \& data mining},
  pages={1269--1278},
  year={2019}
}

@inproceedings{pareja.2020.evolvegcn,
  title={Evolvegcn: Evolving graph convolutional networks for dynamic graphs},
  author={Pareja, Aldo and Domeniconi, Giacomo and Chen, Jie and Ma, Tengfei and Suzumura, Toyotaro and Kanezashi, Hiroki and Kaler, Tim and Schardl, Tao and Leiserson, Charles},
  booktitle={Proceedings of the AAAI conference on artificial intelligence},
  volume={34},
  number={04},
  pages={5363--5370},
  year={2020}
}

@inproceedings{xu.2020.tgat,
  title={Inductive representation learning on temporal graphs},
  author={Xu, Da and Ruan, Chuanwei and Korpeoglu, Evren and Kumar, Sushant and Achan, Kannan},
  booktitle={International Conference on Learning Representations},
  year={2020}
}

@article{cong.2023.graphmixer,
  title={Do we really need complicated model architectures for temporal networks?},
  author={Cong, Weilin and Zhang, Si and Kang, Jian and Yuan, Baichuan and Wu, Hao and Zhou, Xin and Tong, Hanghang and Mahdavi, Mehrdad},
  journal={International Conference on Learning Representations, 2023},
  year={2023}
}

@article{yu.2023.dygformer,
  title={Towards better dynamic graph learning: New architecture and unified library},
  author={Yu, Le and Sun, Leilei and Du, Bowen and Lv, Weifeng},
  journal={Advances in Neural Information Processing Systems},
  volume={36},
  pages={67686--67700},
  year={2023}
}

@inproceedings{zhu.2023.wingnn,
  title={Wingnn: Dynamic graph neural networks with random gradient aggregation window},
  author={Zhu, Yifan and Cong, Fangpeng and Zhang, Dan and Gong, Wenwen and Lin, Qika and Feng, Wenzheng and Dong, Yuxiao and Tang, Jie},
  booktitle={Proceedings of the 29th ACM SIGKDD conference on knowledge discovery and data mining},
  pages={3650--3662},
  year={2023}
}

@inproceedings{beladev.2023.graphert,
author = {Beladev, Moran and Katz, Gilad and Rokach, Lior and Singer, Uriel and Radinsky, Kira},
title = {GraphERT-- Transformers-based Temporal Dynamic Graph Embedding},
year = {2023},
isbn = {9798400701245},
publisher = {Association for Computing Machinery},
address = {New York, NY, USA},
url = {https://doi.org/10.1145/3583780.3614899},
doi = {10.1145/3583780.3614899},
pages = {68–77},
numpages = {10},
location = {Birmingham, United Kingdom},
series = {CIKM '23}
}

@inproceedings{beladev.2020.tdgraphembed,
author = {Beladev, Moran and Rokach, Lior and Katz, Gilad and Guy, Ido and Radinsky, Kira},
title = {tdGraphEmbed: Temporal Dynamic Graph-Level Embedding},
year = {2020},
isbn = {9781450368599},
publisher = {Association for Computing Machinery},
address = {New York, NY, USA},
url = {https://doi.org/10.1145/3340531.3411953},
doi = {10.1145/3340531.3411953},
booktitle = {Proceedings of the 29th ACM International Conference on Information \& Knowledge Management},
pages = {55–64},
numpages = {10},
location = {Virtual Event, Ireland},
series = {CIKM '20}
}

@inproceedings{wang.2023.graph,
author = {Wang, Lili and Huang, Chenghan and Cao, Xinyuan and Ma, Weicheng and Vosoughi, Soroush},
title = {Graph-Level Embedding for Time-Evolving Graphs},
year = {2023},
isbn = {9781450394192},
publisher = {Association for Computing Machinery},
address = {New York, NY, USA},
url = {https://doi.org/10.1145/3543873.3587299},
doi = {10.1145/3543873.3587299},
booktitle = {Companion Proceedings of the ACM Web Conference 2023},
pages = {5–8},
numpages = {4},
location = {Austin, TX, USA},
series = {WWW '23 Companion}
}

@article{melnyk.2023.understanding,
  title={Understanding microbiome dynamics via interpretable graph representation learning},
  author={Melnyk, Kateryna and Weimann, Kuba and Conrad, Tim OF},
  journal={Scientific Reports},
  volume={13},
  number={1},
  pages={2058},
  year={2023},
  publisher={Nature Publishing Group UK London}
}

@ARTICLE{feng.2026.survey,
  author={Feng, ZhengZhao and Wang, Rui and Wang, TianXing and Song, Mingli and Wu, Sai and He, Shuibing},
  journal={IEEE Transactions on Knowledge and Data Engineering}, 
  title={A Comprehensive Survey of Dynamic Graph Neural Networks: Models, Frameworks, Benchmarks, Experiments and Challenges}, 
  year={2026},
  volume={38},
  number={1},
  pages={26-46},
  doi={10.1109/TKDE.2025.3621291}
}

@article{jia.2020.dstgnn,
  title={Dynamic spatiotemporal graph neural network with tensor network},
  author={Jia, Chengcheng and Wu, Bo and Zhang, Xiao-Ping},
  journal={arXiv preprint arXiv:2003.08729},
  year={2020}
}

@article{zhao.2023.rtgnn,
author = {Zhao, Xusheng and Dai, Qiong and Wu, Jia and Peng, Hao and Liu, Mingsheng and Bai, Xu and Tan, Jianlong and Wang, Senzhang and Yu, Philip S.},
title = {Multi-View Tensor Graph Neural Networks Through Reinforced Aggregation},
year = {2023},
issue_date = {April 2023},
publisher = {IEEE Educational Activities Department},
address = {USA},
volume = {35},
number = {4},
issn = {1041-4347},
url = {https://doi.org/10.1109/TKDE.2022.3142179},
doi = {10.1109/TKDE.2022.3142179},
journal = {IEEE Trans. on Knowl. and Data Eng.},
month = apr,
pages = {4077–4091},
numpages = {15}
}

@article{zheng.2025.mantae,
  title={Mode-Aware Non-Linear Tucker Autoencoder for Tensor-based Unsupervised Learning},
  author={Zheng, Junjing and Song, Chengliang and Jiang, Weidong and Zhang, Xinyu},
  journal={arXiv preprint arXiv:2508.06784},
  year={2025}
}

@article{xie.2023.datc,
author = {Xie, Kun and Ouyang, Yudian and Wang, Xin and Xie, Gaogang and Li, Kenli and Liang, Wei and Cao, Jiannong and Wen, Jigang},
title = {Deep Adversarial Tensor Completion for Accurate Network Traffic Measurement},
year = {2023},
issue_date = {Oct. 2023},
publisher = {IEEE Press},
volume = {31},
number = {5},
issn = {1063-6692},
url = {https://doi.org/10.1109/TNET.2022.3233908},
doi = {10.1109/TNET.2022.3233908},
journal = {IEEE/ACM Trans. Netw.},
month = feb,
pages = {2101–2116},
numpages = {16}
}

@article{zhao.2026.deep,
title = {Deep tensor decomposition: A survey},
journal = {Neurocomputing},
volume = {664},
pages = {132074},
year = {2026},
issn = {0925-2312},
doi = {https://doi.org/10.1016/j.neucom.2025.132074},
url = {https://www.sciencedirect.com/science/article/pii/S0925231225027468},
author = {Meng Zhao and Jiuyun Hu and Ziyue Li and Zekai Wang and Lijun Sun},
}

@article{han.2015.epidemic,
title = {Epidemic process on activity-driven modular networks},
journal = {Physica A: Statistical Mechanics and its Applications},
volume = {432},
pages = {354-362},
year = {2015},
issn = {0378-4371},
doi = {https://doi.org/10.1016/j.physa.2015.03.062},
url = {https://www.sciencedirect.com/science/article/pii/S0378437115003246},
author = {Dun Han and Mei Sun and Dandan Li},
}

@inproceedings{kumar.2019.predicting,
  title={Predicting dynamic embedding trajectory in temporal interaction networks},
  author={Kumar, Srijan and Zhang, Xikun and Leskovec, Jure},
  booktitle={Proceedings of the 25th ACM SIGKDD international conference on knowledge discovery \& data mining},
  pages={1269--1278},
  year={2019}
}

@article{shetty.2004.enron,
  title={The Enron email dataset database schema and brief statistical report},
  author={Shetty, Jitesh and Adibi, Jafar},
  journal={Information sciences institute technical report, University of Southern California},
  volume={4},
  number={1},
  pages={120--128},
  year={2004},
  publisher={Citeseer}
}
